\documentclass{article}

\usepackage[utf8]{inputenc}
\usepackage[T1]{fontenc}
\usepackage{url}            
\usepackage{amsfonts}       
\usepackage{nicefrac}       
\usepackage{subcaption}
\usepackage{amsmath}
\usepackage{bm}             
\usepackage{subcaption}
\usepackage{graphicx}
\usepackage{subcaption}     
\usepackage{adjustbox}      
\graphicspath{{media/}}     

\usepackage{booktabs}       
\usepackage{multirow}       
\usepackage{colortbl}       
\usepackage{xcolor}         

\usepackage{algorithm}
\usepackage{algorithmic}

\usepackage{listings}
\usepackage{amsthm}
\usepackage{fancyhdr}       
\usepackage{microtype}      
\usepackage{placeins}       

\usepackage{PRIMEarxiv}
\usepackage{lipsum}         

\usepackage{hyperref}
\usepackage{cleveref}
\title{Efficient Diffusion Distillation via Embedding Loss}

\author{
  Jincheng Ying\thanks{Equal contribution.},
  Yitao Chen\protect\footnotemark[1],
  Li Wenlin \\
  School of Big Data and Artificial Intelligence \\
  Guangdong University of Finance and Economics \\
  Guangzhou, China \\
  \texttt{jc\_ying@student.gdufe.edu.cn},
  \texttt{teochen@student.gdufe.edu.cn} \\
  \texttt{wenlin@gdufe.edu.cn} \\          
  \And
  Minghui Xu \\
  School of Computer Science \\
  Shandong University \\
  Qingdao, China \\
  \texttt{mhxu@sdu.edu.cn} \\
  \And
  Yinhao Xiao\thanks{Corresponding author.} \\
  School of Big Data and Artificial Intelligence \\
  Guangdong University of Finance and Economics \\
  Guangzhou, China \\
  \texttt{20191081@gdufe.edu.cn} \\
}

\begin{document}
\maketitle

\begin{abstract}
Recent advances in distilling expensive diffusion models into efficient few-step generators show significant promise. However, these methods typically demand substantial computational resources and extended training periods, limiting accessibility for resource-constrained researchers, and existing supplementary loss functions have notable limitations. Regression loss requires pre-generating large datasets before training and limits the student model to the teacher's performance, while GAN-based losses suffer from training instability and require careful tuning. In this paper, we propose Embedding Loss (EL), a novel supplementary loss function that complements existing diffusion distillation methods to enhance generation quality and accelerate training with smaller batch sizes. Leveraging feature embeddings from a diverse set of randomly initialized networks, EL effectively aligns the feature distributions between the distilled few-step generator and the original data. By computing Maximum Mean Discrepancy (MMD) in the embedded feature space, EL ensures robust distribution matching, thereby preserving sample fidelity and diversity during distillation. Within distribution matching distillation frameworks, EL demonstrates strong empirical performance for one-step generators. On the CIFAR-10 dataset, our approach achieves state-of-the-art Fréchet Inception Distance (FID) values of 1.475 for unconditional generation and 1.380 for conditional generation. Beyond CIFAR-10, we further validate EL across multiple benchmarks and distillation methods, including ImageNet (64×64 and 512×512), AFHQ-v2, and FFHQ datasets, using DMD, DI, and CM distillation frameworks, demonstrating consistent improvements over existing one-step distillation methods. Our method also reduces training iterations by up to 80\%, offering a more practical and scalable solution for deploying diffusion-based generative models in resource-constrained environments.

\end{abstract}


\section{Introduction}
\label{intro}

Diffusion models have emerged as the leading approach for high-quality image generation \cite{hoDenoisingDiffusionProbabilistic2020a,song2019generative}. These models demonstrate superior training stability, robustness against mode collapse, and the ability to generate diverse photorealistic images \cite{beatsGAN,ho_intro,ramesh2022_intro}. However, their iterative sampling process requires numerous forward passes through the generative network, typically ranging from 50 to 1000 steps, resulting in slow inference that limits their practical deployment in real-world applications.

To address this limitation, various distillation methods have been proposed to compress expensive diffusion models into efficient few-step generators. These approaches, including Progressive Distillation \cite{salimansProgressiveDistillationFast2022}, Consistency Distillation \cite{songConsistencyModels2023}, and distribution matching frameworks \cite{luoDiffInstructUniversalApproach2023,yinOnestepDiffusionDistribution2024,SID,SIM}, have achieved strong results in reducing sampling steps while maintaining generation quality. However, they typically require extensive training with large batch sizes to achieve competitive performance. For instance, Consistency Distillation (CD) \cite{songConsistencyModels2023} requires 800K iterations with a batch size of 512 on CIFAR-10, while DMD \cite{yinOnestepDiffusionDistribution2024} requires 350K iterations with a batch size of 392 on ImageNet. Such computational demands, often requiring days to weeks of multi-GPU training, pose significant challenges in resource-constrained settings. When limited hardware necessitates smaller batch sizes, training suffers from slower convergence and reduced generalization, while extended training times further limit practical applicability.

Existing distillation methods often incorporate supplementary loss functions to improve generation quality. However, current approaches have notable limitations that restrict their effectiveness and accessibility. Regression-based losses such as DMD \cite{yinOnestepDiffusionDistribution2024}, which directly minimize pixel-space differences between student and teacher outputs, require pre-generating and storing large datasets before training commences. This requires significant storage resources and fundamentally limits the student model's performance to at most match the teacher model, creating a performance ceiling. Alternatively, GAN-based losses \cite{goodfellow2020generative,yinImprovedDistributionMatching2024} have been employed to enhance sample quality through adversarial training. While potentially effective, these methods introduce significant training instability and mode collapse risks, requiring careful hyperparameter tuning and often demanding even larger batch sizes to stabilize the adversarial dynamics. Furthermore, GAN-based approaches add substantial computational overhead through the discriminator network and its optimization. These limitations motivate the need for a supplementary loss function that is both stable to train and computationally efficient, while avoiding the performance ceiling imposed by regression losses.

In this study, we analyze why existing methods require large batches and propose Embedding Loss (EL), a novel supplementary loss function that enhances both the quality and training efficiency of diffusion model distillation with smaller batch sizes, without introducing excessive computational or memory overhead. EL aligns the feature distributions between the distilled one-step generator and real data through an ensemble of randomly initialized neural networks with diverse architectures. By measuring Maximum Mean Discrepancy (MMD) \cite{MMD} in the embedded feature space, EL addresses this approximation problem and ensures robust distribution matching, thereby preserving both fidelity and diversity in generated samples with smaller batch sizes and fewer iterations.

Our approach is broadly applicable and computationally efficient. By integrating EL into existing distribution matching frameworks such as DI \cite{luoDiffInstructUniversalApproach2023}, SiD\textsuperscript{2}A \cite{SID2}, and DMD \cite{yinOnestepDiffusionDistribution2024}, we demonstrate substantial gains: EL enables one-step generators to achieve state-of-the-art Fréchet Inception Distance (FID) scores \cite{FID} of 1.475 for unconditional and 1.380 for conditional generation on CIFAR-10 \cite{cifar10}, a significant advance in fast generative modeling. Moreover, EL delivers consistent improvements across multiple benchmarks, including AFHQ-v2 \cite{AFHQ}, ImageNet 64×64 \cite{imagenet}, and FFHQ \cite{FFHQ}, outperforming prior methods by considerable margins.

Crucially, our method reduces the required training iterations by up to 80\%, significantly streamlining the deployment of diffusion-based generative models in resource-constrained settings. Our implementation is available at https://github.com/hahahaj123/EL

Our main contributions are as follows:

\begin{itemize}
\item We analyze why existing distribution matching methods require large batch sizes for effective training, identifying an approximation gap that impacts both training efficiency and sample quality when using smaller batches.

\item We propose Embedding Loss (EL), a novel auxiliary loss that measures feature-space distribution discrepancy via Maximum Mean Discrepancy (MMD) using an ensemble of randomly initialized neural networks. EL reduces required training iterations by up to 80\% and enables effective distillation with smaller batch sizes, without significant computational overhead. It can be seamlessly integrated into various distribution matching frameworks (DI, SiD\textsuperscript{2}A, DMD).

\item We achieve state-of-the-art results for one-step generation across multiple benchmarks: FID scores of 1.475 (unconditional) and 1.380 (conditional) on CIFAR-10, with consistent improvements on AFHQ-v2, ImageNet $64\times64$, $512\times512$, and FFHQ.
\end{itemize}

\section{Related Work}

\subsection*{Diffusion Acceleration}
Substantial research has focused on accelerating the reverse diffusion process to reduce the number of sampling steps required. One major approach reformulates the stochastic differential equation (SDE) into an ordinary differential equation (ODE), which enables deterministic sampling \cite{ODE,lipman2022flow_matching,karrasElucidatingDesignSpace2022,luDPMSolverFastODE2022,zhang_chen2022fast}. Despite these advances, a notable trade-off persists between reducing sampling steps and maintaining visual quality.

Another research direction considers diffusion models within the flow matching framework, employing strategies to transform the reverse diffusion process into more linear trajectories, thus enabling larger step reductions \cite{lipman2022flow_matching,liu2022flow}. To achieve generation in fewer steps, researchers have also proposed truncating the diffusion chain and initiating generation from an implicit distribution instead of white Gaussian noise \cite{pandey2022diffusevae,lyu2022accelerating}, as well as combining diffusion models with GANs to enable faster generation \cite{wang2022diffusion,xiao2021tackling}.

\subsection*{Existing Diffusion Distillation Frameworks}

Current diffusion distillation frameworks fall into three main categories \cite{renHyperSDTrajectorySegmented2024}: trajectory-preserving distillation, trajectory-reformulating distillation, and distribution-matching distillation.

\subsubsection*{Trajectory-Preserving Distillation}
Trajectory-preserving methods aim to maintain the solution trajectory defined by the ordinary differential equation (ODE) of the diffusion process. Representative works such as Progressive Distillation~\cite{salimansProgressiveDistillationFast2022} and Consistency Models~\cite{songConsistencyModels2023} ensure that the student model's outputs closely match those of the teacher model throughout the sampling trajectory. By enabling the student to directly predict intermediate states, these methods reduce inference steps while effectively replicating the teacher's state transitions. Some implementations additionally incorporate adversarial losses~\cite{lin2024sdxl_lightning} to enhance distributional similarity and output fidelity. However, these approaches are fundamentally constrained by ODE fitting accuracy, as approximation errors can accumulate and degrade sample quality, especially under aggressive distillation.

\subsubsection*{Trajectory-Reformulating Distillation}
Trajectory-reformulating methods, such as ADD~\cite{sauerAdversarialDiffusionDistillation2023}, Rectified Flow~\cite{Rectified_Flow}, and LADD~\cite{LADD}, directly leverage the ODE trajectory endpoints or real images as primary supervision, bypassing intermediate steps of the original trajectory. By constructing more efficient pathways, these methods achieve further reductions in inference steps. Releasing the student model from strict adherence to the teacher's trajectory enables trajectory-reformulating distillation to achieve greater flexibility in few-step generation. However, this flexibility may introduce inconsistencies between the distilled model's outputs and the original teacher model, occasionally leading to undesired or unstable generation outcomes.

\subsubsection*{Distribution-Matching Distillation}
Distribution-matching distillation, also known as Score Distillation Sampling (SDS), is a framework first proposed by Diff-Instruct (DI)~\cite{luoDiffInstructUniversalApproach2023}, with subsequent methods including SiD~\cite{SID}, DMD~\cite{yinOnestepDiffusionDistribution2024}, and SIM~\cite{SIM}. This approach utilizes $f_{\text{teacher}}$ to estimate the score for the real distribution and introduces a fake score model $f_{\text{fake}}$ to estimate the score for the fake distribution (i.e., the student model's distribution), thereby enabling one-step inference.

To improve model performance, recent distribution-matching methods have introduced auxiliary losses. DMD~\cite{yinOnestepDiffusionDistribution2024} incorporates regression loss by pre-generating target images using the teacher model, though this incurs substantial computational overhead and limits student performance to the teacher's capabilities. DMD2~\cite{yinImprovedDistributionMatching2024} adopts adversarial loss through GANs to avoid pre-generation, but introduces the complexities inherent in adversarial training~\cite{sauerAdversarialDiffusionDistillation2023}, including sensitivity to hyperparameters, careful balancing between discriminator and generator, and risk of mode collapse or divergence.

All the distillation methods mentioned above share a common limitation: they typically demand substantial computational resources and extended training periods.

To address these computational challenges, we propose Embedding Loss (EL), a novel auxiliary loss that significantly reduces both resource requirements and training time while addressing the inherent shortcomings of existing auxiliary losses. Unlike regression loss, EL eliminates the need to pre-generate large target datasets, thereby removing considerable computational overhead while enabling the student model to potentially exceed teacher performance through direct alignment with real data distributions. In contrast to GAN-based losses, EL circumvents the training instability associated with adversarial objectives by employing Maximum Mean Discrepancy (MMD) in randomly initialized embedding spaces, a stable, non-adversarial metric that demands minimal hyperparameter tuning. Moreover, EL facilitates effective training with smaller batch sizes, reducing memory requirements and making high-quality distillation accessible to researchers with constrained computational budgets. By computing distributional distances in diverse feature spaces rather than pixel space, EL delivers robust supervision that complements the score distillation objective while preserving training stability across different distillation frameworks.

\begin{figure*}[t]
    \centering
    \includegraphics[width=1.0\textwidth]{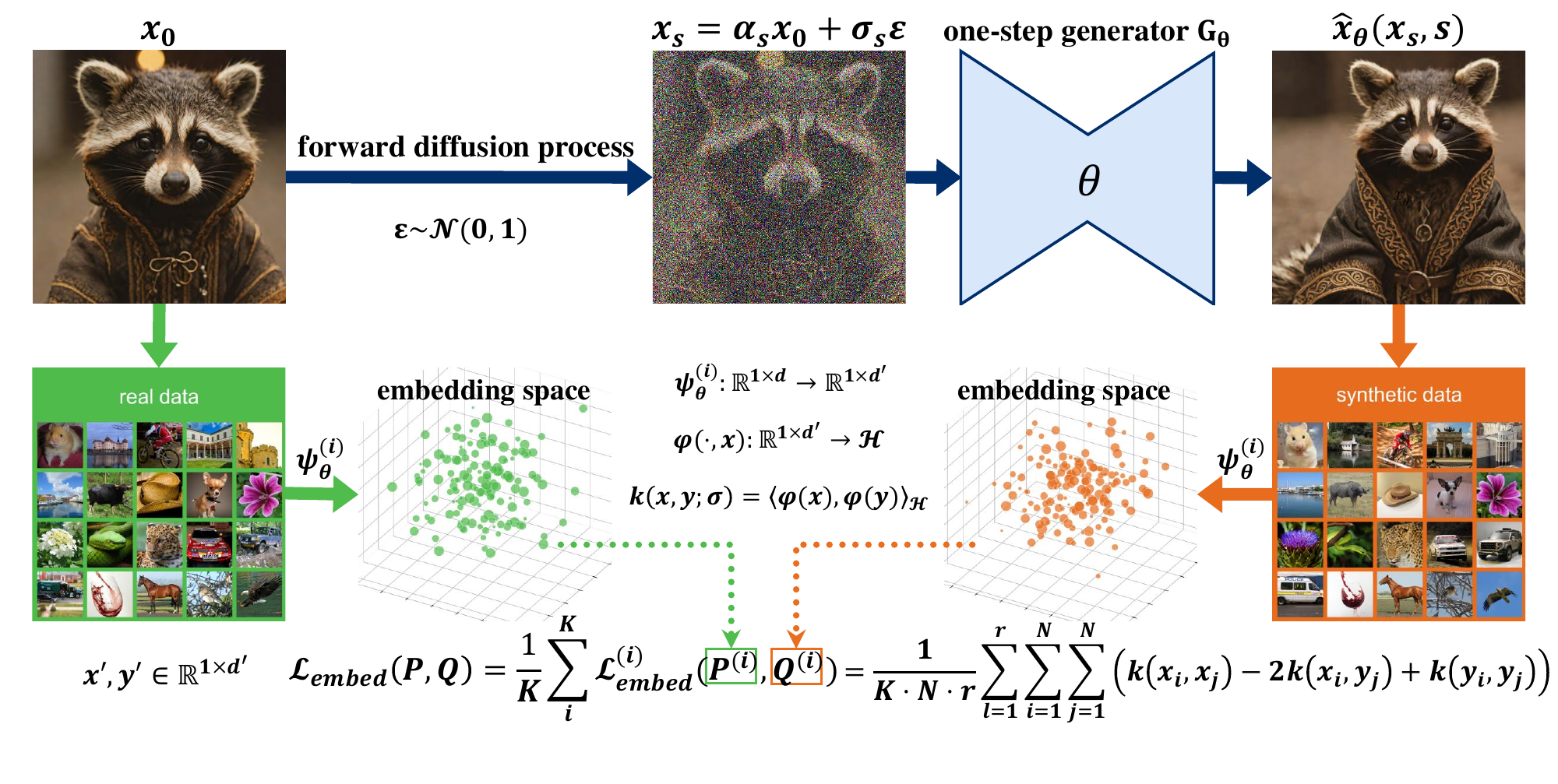}
    \caption{\textbf{Method overview.} We train a one-step generator $\bm{G_\theta}$ to map noisy images into realistic outputs while maintaining distributional alignment with real data. The framework consists of three key components: \textbf{(1) Forward diffusion and denoising pipeline (top row).} Clean images $\bm{x}_0$ (e.g., the raccoon portrait) undergo forward diffusion by adding Gaussian noise $\boldsymbol{\epsilon} \sim \mathcal{N}(\mathbf{0}, \mathbf{I})$ to produce noisy images $\bm{x_s} = \bm{\alpha_s} \bm{x_0} + \bm{\sigma_s} \boldsymbol{\epsilon}$. The one-step generator $\bm{G_\theta}$ then denoises these to produce clean synthetic samples $\hat{\bm{x}}_0(\bm{x_s}, \bm{s})$. \textbf{(2) Embedding space alignment using Maximum Mean Discrepancy (bottom center).} Real data samples are embedded via $\bm{\psi_\theta}^{(i)}$, which reduces dimensionality from $1 \times d$ to $1 \times d'$, and $\bm{\phi}(\cdot, \bm{x})$ into a reproducing kernel Hilbert space $\bm{\mathcal{H}}$ (green dots), while synthetic samples are similarly embedded (orange dots). We draw $N$ samples from each embedding space to compute the MMD statistic. The MMD loss $\bm{\mathcal{L}_{\bm{\textit{embed}}}(\bm{P}, \bm{Q})}$ minimizes the distributional distance using $\bm{K}$ embedding functions $\bm{\psi^{(i)}}$ and $\bm{r}$ Gaussian kernel functions $k(\bm{x}, \bm{y}; \bm{\sigma^{(j)}})$, ensuring generated samples are statistically indistinguishable from real data. \textbf{(3) Real and synthetic data distributions (bottom panels).} Real training data (green panel, left) includes diverse semantic categories such as cars, flowers, landscapes, animals, and buildings. The generator produces corresponding high-quality synthetic augmentations (orange panel, right) that preserve the statistical properties and visual fidelity of the original data distribution. This approach enables efficient one-step generation and data augmentation while maintaining distributional alignment.}
    \label{fig:embedding_loss_full}
\end{figure*}

\section{Distribution Matching through Embedding Loss}

Diffusion distillation aims to compress multi-step pre-trained diffusion models into efficient few-step generators that produce high-quality images without expensive iterative sampling. While existing distillation methods have demonstrated promising results, they face a critical bottleneck in practical deployment. As we detail in Section~\ref{sec:3.1}, achieving satisfactory performance requires extremely long training times and prohibitively large batch sizes, severely limiting applicability when hardware resources are constrained.

To address these challenges, we conduct a theoretical analysis of the distribution matching distillation framework and identify the fundamental source of its training inefficiency (Section~\ref{sec:3.2}). Our analysis reveals that small batch sizes lead to high gradient variance due to poor approximation of the data distribution and multiple independent variance sources beyond standard Monte Carlo variance. While existing methods employ auxiliary losses to mitigate this issue, they introduce significant drawbacks (Section~\ref{sec:Limitations of Existing Auxiliary Losses}): regression losses require expensive offline dataset pre-generation and suffer from dataset staleness, while adversarial losses create training instability through non-stationary optimization and add substantial computational overhead ($1.5\times$).

Inspired by recent advances in dataset condensation \cite{zhaoDatasetCondensationDistribution2023}, we propose an embedding-based loss that aligns feature distributions between real and generated images using Maximum Mean Discrepancy (MMD) \cite{MMD} computed in randomly projected embedding spaces. Unlike existing auxiliary losses, our approach requires no pre-computation, introduces minimal computational overhead, and maintains training stability through frozen feature extractors. We provide theoretical analysis demonstrating that the embedding loss effectively reduces gradient variance and accelerates convergence when combined with the distribution matching objective (Section~\ref{sec:3.2.4}).

Adding the proposed loss to the Distribution Matching framework achieves better generation quality than the original while significantly reducing training time. Experiments also show that it works well in trajectory-preserving methods such as Consistency Distillation~\cite{songConsistencyModels2023} (Section~\ref{Generality of Embedding Loss}).

\subsection{Diffusion Distillation Problem}
\label{sec:3.1}
Diffusion distillation aims to compress multi-step diffusion models into efficient few-step generators that retain strong generation quality. The goal is to accelerate the costly iterative sampling of pre-trained models by training student models to synthesize high-quality samples in fewer steps. 
Despite achieving competitive results, existing methods face a major challenge: they require excessive training iterations and large batch sizes. For instance, Consistency Distillation (CD) and Consistency Training (CT) \cite{songConsistencyModels2023} need over 800K iterations with batch size 512, while DMD \cite{yinOnestepDiffusionDistribution2024} requires around 350K iterations with batch size 392. 
As shown in Table \ref{CM:cifar10_comparison}, small batch sizes and fewer iterations can degrade performance. These requirements make deployment difficult with limited hardware.

\subsection{Theoretical Analysis}
\label{sec:3.2}

\subsubsection{Problem Setup}
We denote by $\phi$ and $\theta$ the parameters of teacher and student networks, with corresponding score functions $s_{\phi}$ and $s_{\theta}$. Let $G_{\theta}$ denote the student generator, $x_t$ the noisy sample at timestep $t$, and $p_t(x_t)$ the marginal distribution.

\textbf{Assumption 1 (Well-trained Teacher):} 
The teacher satisfies $\mathbb{E}_{x_t \sim p_{\text{data}}(x_t|t)}[ \| s_{\phi}(x_t, t) - \nabla_{x_t} \log p_t(x_t) \|^2 ] \leq \epsilon^2(t)$.

\subsubsection{The Batch Size Challenge in Distribution Matching}

Distribution matching distillation minimizes:
\begin{equation}
\mathcal{L}_{\text{DM}}(\theta) = \int_{t=0}^{T} w(t) \mathbb{E}_{z, x_0, x_t} \| s_{\theta}(x_t, t) - s_{\phi}(x_t,t) \|^2 dt
\end{equation}

\textbf{Proposition 1:} Under Assumption 1, at local minimum $\theta^*$:
\begin{equation}
D_{KL}(p_{\text{data}} \| p_{\theta^*}) \leq C_1 \epsilon_{\text{teacher}}^2 + C_2 \epsilon_{\text{opt}}
\end{equation}
where $\epsilon_{\text{opt}}$ captures the optimization error.

In practice, small batch sizes lead to high optimization error because:
\begin{enumerate}
\item \textbf{High gradient variance}: The student matches an empirical distribution $\hat{p}_{\text{data}}$ poorly approximating $p_{\text{data}}$.
\item \textbf{Multiple variance sources}: Beyond standard Monte Carlo variance $\mathcal{O}(1/B)$, there exist batch-independent variances from noise injection ($\sigma^2_{\text{noise}}$), timestep sampling ($\sigma^2_{\text{time}}$), and teacher approximation ($\sigma^2_{\text{diffusion}}$)
\end{enumerate}.

This explains why existing methods require large batches (336 in DMD, 2048 in CD). See Appendix~\ref{app:variance_analysis} for detailed variance decomposition.

\subsubsection{Limitations of Existing Auxiliary Losses}
\label{sec:Limitations of Existing Auxiliary Losses}

To address gradient variance, prior work employs auxiliary losses:

\textbf{Regression loss} \cite{yinOnestepDiffusionDistribution2024} minimizes $\mathcal{L}_{\text{reg}}(\theta) = \mathbb{E}_{(z,y) \sim \mathcal{D}}[\ell(G_\theta(z), y)]$ where $\mathcal{D}$ contains pre-generated teacher outputs. This requires expensive offline generation ($\sim$500k pairs) and suffers from dataset staleness.

\textbf{Adversarial loss} \cite{yinImprovedDistributionMatching2024} trains a discriminator $D$ alongside the generator:
\begin{equation}
\mathcal{L}_{\text{adv}}(\theta) = \mathbb{E}_{z, t}[-\log D(F(G_\theta(z), t))]
\end{equation}
This introduces non-stationary optimization, gradient instability, and $1.5\times$ computational overhead.

\subsubsection{Embedding Loss as Superior Alternative}
\label{sec:3.2.4}

We propose multi-scale embedding loss using frozen feature extractors $\{\psi_i\}_{i=1}^K$:
\begin{equation}
\mathcal{L}_{\text{embed}}(\theta) = \frac{1}{K} \sum_{i=1}^K D_{\text{MMD}^2}(p_{\text{data}}, p_\theta;\psi_i)
\end{equation}

\textbf{Theorem 1 (Key Properties):} The embedding loss gradient:
\begin{itemize}
\item Decomposes into alignment (with real data) and diversity (among generations) terms
\item Has bounded variance: $\text{Var}(\nabla_\theta \mathcal{L}_{\text{Embed}}) \leq O(1/B) + O(1/M)$
\item Remains stable with frozen extractors (no gradient explosion)
\end{itemize}

\textbf{Practical advantages:}
\begin{table}[h]
\centering
\small
\begin{tabular}{lccc}
\toprule
Method & Pre-compute & Comp. Cost & Stability \\
\midrule
Regression & Yes ($\sim$500k) & Mid & Moderate \\
Adversarial & No & High ($1.5\times$) & Low \\
\textbf{Embed (Ours)} & No & Low & High \\
\bottomrule
\end{tabular}
\caption{Comparison of auxiliary losses}
\label{tab:aux_comparison}
\end{table}

\textbf{Theorem 2 (Convergence):} Combining $\mathcal{L}_{\text{total}} = (1-\lambda)\mathcal{L}_{\text{DM}} + \lambda \mathcal{L}_{\text{embed}}$ achieves faster convergence when gradients are positively correlated ($\rho > 0$), with optimal $\lambda^*$
\begin{equation*}
    \lambda^* =   \frac{\sigma_{\text{DM}}^2 - \rho \sigma_{\text{DM}} \sigma_{\text{embed}}}{\sigma_{\text{DM}}^2 + \sigma_{\text{embed}}^2 - 2 \rho \sigma_{\text{DM}} \sigma_{\text{embed}}}.
\end{equation*}
At this optimal $\lambda^*$, the minimal gradient variance is achieved as $\text{Var}(g^*_{\text{total}}) = \frac{\sigma_1^2 \sigma_2^2 (1 - \rho^2)}{\sigma_1^2 + \sigma_2^2 - 2\rho \sigma_1 \sigma_2}$

Complete proofs and derivations are in Appendix~\ref{app:Variance_Reduction_with_Positive_Correlation}.

\subsection*{Architecture Diversity}
	
	\textbf{Motivation.} While the original dataset condensation method uses a single ConvNet architecture for embedding, we find this insufficient for diffusion distillation (see Section~\ref{sec:ablation}). This limitation stems from the high-dimensional nature of the distribution matching objective in $\mathcal{L}_{\text{DM}}$. 
	
	\textbf{Theoretical Justification.} The gradient variance in distribution matching satisfies:
	\begin{equation}
		\text{Var}(\widehat{\text{Grad}}(\theta)) = \sigma^2_{\text{noise}} + \sigma^2_{\text{time}} + \sigma^2_{\text{diffusion}} + \mathcal{O}(1/B)
	\end{equation}
	
	When batch size $B$ is small, the $\mathcal{O}(1/B)$ term dominates, making gradient estimation highly unstable.Complete proofs and derivations are in Appendix~\ref{app:variance_analysis}. To compensate for this without increasing $B$, we need to reduce the other variance components. The key insight is that using diverse embedding architectures $\{\psi_k\}_{k=1}^K$ effectively provides multiple independent views of the same distribution, which helps in two ways:
	
	\begin{enumerate}
		\item \textbf{Variance reduction through averaging}: With $K$ diverse embeddings, the effective gradient becomes:
		\begin{equation}
			\widehat{\text{Grad}}_{\text{ensemble}}(\theta) = \frac{1}{K}\sum_{k=1}^K \widehat{\text{Grad}}_k(\theta)
		\end{equation}
		If in the case of $K$ embeddings, these embeddings capture complementary aspects of the distribution, the variance is reduced by approximately $\mathcal{O}(1/K)$ through ensemble averaging.
		
		\item \textbf{Improved distributional coverage}: A single architecture $\psi$ may have blind spots, regions of the data distribution where its embedding $\psi(x)$ provides poor discriminative power. Using diverse architectures ensures that at least some $\psi_k$ will provide informative gradients in any region, preventing mode collapse when $B$ is small.
	\end{enumerate}


\textbf{Architecture Design.} To address this, we employ a diverse set of embedding architectures:
\begin{enumerate}
    \item \textbf{Simple CNN}: Progressive pooling for efficient low-level feature extraction, capturing basic spatial patterns
    \item \textbf{Multi-scale network}: Captures features at different spatial resolutions, providing scale-invariant representations crucial for matching scores across different diffusion timesteps $t$
    \item \textbf{Residual network}: Enables deeper feature learning through skip connections, addressing the temporal dynamics in $\int_{t=0}^T w(t)\mathbb{E}[\cdot]dt$
    \item \textbf{Attention-based network}: Adaptively weights spatial features, focusing on discriminative regions that matter most for score matching
\end{enumerate}

Each architecture is initialized using different strategies (Xavier, Kaiming, normal, and orthogonal initialization), further increasing the diversity of learned representations. This initialization diversity ensures that the networks explore different regions of the parameter space, leading to complementary feature extractors.

All networks map inputs to a common $d'$-dimensional embedding space ($d' = 64$ or $128$ in our experiments), are frozen during training (i.e., $\partial \psi_k / \partial \theta = 0$), and incur negligible computational overhead due to their lightweight design. The frozen weights are critical, they prevent the embeddings from collapsing to trivial solutions and maintain the diversity throughout training.
Algorithm \ref{alg:embedding-loss} describes the ultimate training process.

\begin{algorithm}[t]
\caption{Embedding Loss with Multiple Random Networks}
\label{alg:embedding-loss}
\begin{algorithmic}[1]
\REQUIRE $G_{imgs}$, $R_{imgs}$, $M=1$, $d=64/128$, $\lambda_{embed}$
\ENSURE $\mathcal{L}_{total} / (K \times M)$
   \STATE $\mathcal{L}_{total} \gets 0$
   \STATE $\mathcal{T} \gets \{\text{SimpleCNN, MultiScale, Residual, Attention}\}$
   \STATE $\mathcal{S} \gets \{\text{Xavier, Kaiming, Normal, Orthogonal}\}$
   \STATE $K \gets |\mathcal{T}|$
   \FOR{$i = 1$ to $K$}
       \FOR{$j = 1$ to $M$}  
           \STATE $\tau \gets \mathcal{T}[i]$ 
           \STATE $s \gets \mathcal{S}[i]$ 
           \STATE $\mathcal{F}_i \gets \text{CreateNetwork}(\tau, d)$
           \STATE $\text{Initialize}(\mathcal{F}_i, s)$
           \STATE $\mathcal{F}_i.\text{eval}()$
           \FOR{$p \in f_\theta.\text{parameters}()$}
               \STATE $p.\text{requires\_grad} \gets \text{False}$
           \ENDFOR
           \STATE $\mathbf{z}_r \gets \mathcal{F}_i(R_{imgs})$
           \STATE $\mathbf{z}_g \gets \mathcal{F}_i(G_{imgs})$
           \STATE $\mathcal{L}_{total} \gets \mathcal{L}_{total} + \lambda_{embed} \cdot D_{\textit{MMD}^2}(\mathbf{z}_r, \mathbf{z}_g)$
       \ENDFOR
   \ENDFOR
   \STATE \textbf{return} $\mathcal{L}_{total} / (K \times M)$
\end{algorithmic}
\end{algorithm}

\textbf{Practical Impact.} This architectural diversity acts as an implicit regularizer that stabilizes training with small batches ($B \ll 100$), reducing the effective variance without the computational burden of large batches. Our ablation studies (Section~\ref{sec:ablation}) show that removing architectural diversity degrades performance significantly, confirming its importance for compensating the $\mathcal{O}(1/B)$ variance term.

\subsection{Generality of Embedding Loss}
\label{Generality of Embedding Loss}
EL extends beyond DMD\cite{yinOnestepDiffusionDistribution2024} to other distillation methods. Tables \ref{tab:cifar10_uncon}, \ref{tab:cifar10_cond} and \ref{CM:cifar10_comparison} shows that adding EL to Consistency Distillation (CD)\cite{songConsistencyModels2023}, Score Identity Distillation (SiD)\cite{SID,SID2} and DI\cite{luoDiffInstructUniversalApproach2023}also improves performance. This works because EL provides a distribution-level training signal that is independent of the specific distillation objective, whether trajectory-preserving (CD) or distribution-matching (DMD, SiD,DI). By aligning generated samples with real data in embedding space, EL complements any distillation method that produces a one-step or few-step generator.

\begin{table*}[t]
\centering
\begin{minipage}[t]{0.48\textwidth}
\centering
\caption{Comparison of unconditional generation on CIFAR-10. The best one/few-step generator under the FID or IS metric is highlighted with \textbf{bold}.}
\label{tab:cifar10_uncon}
\footnotesize
\adjustbox{max width=\textwidth}{%
\begin{tabular}{llccc}
\toprule
Family & Model & NFE & FID ($\downarrow$) & IS ($\uparrow$) \\
\midrule
Teacher & VP-EDM\cite{karrasElucidatingDesignSpace2022}& 35 & 1.97 & 9.68 \\
\midrule
\multirow{8}{*}{Diffusion} 
& DDPM\cite{DDPM} & 1000 & 3.17 & 9.46 \\
& DDIM\cite{DDIM} & 100 & 4.16 & \\
& DPM-Solver-3\cite{lu2022dpm_solver} & 48 & 2.65 & \\
& VDM\cite{VDM} & 1000 & 4.00 & \\
& iDDPM\cite{iddpm} & 4000 & 2.90 & \\
& HSIvI-SM\cite{yu2023hierarchical} & 15 & 4.17 & \\
& VP-EDM+LEGO-PR\cite{vp+edm+lego+pr}& 35 & 1.88 & 9.84 \\
\midrule
\multirow{16}{*}{One Step} 
&StyleGAN2+ADA+Tune\cite{StyleGAN2+ADA+Tune} & 1 & 2.92 & 9.83 \\
&Diffusion ProjectedGAN\cite{DiffusionProjectedGAN}& 1 & 2.54 & \\
& iCT-deep\cite{song2023improvedCM} & 1 & 2.51 & 9.76 \\
& StyleGAN2+ADA+Tune\cite{luoDiffInstructUniversalApproach2023}& 1 & 2.71 & 9.86 \\
& DMD\cite{yinOnestepDiffusionDistribution2024} & 1 & 3.77 & \\
& Diff-Instruct
\cite{luoDiffInstructUniversalApproach2023}& 1 & 4.53 & \\
& \textbf{Diff-Instruct + EL (ours)} & 1 & 3.95 & \\

& CTM\cite{CTM} & 1 & 1.98 & \\
& GDD-I\cite{GDD}& 1 & 1.54 & 10.10 \\
& SiD, $\alpha = 1.0$\cite{SID} & 1 & 2.028 & 10.017 \\
& SiD, $\alpha = 1.2$\cite{SID} & 1 & 1.923 & 9.980 \\
& SiDA, $\alpha = 1.0$\cite{SID2} & 1 & 1.516 & \textbf{10.323} \\
& SiD$^2$A, $\alpha = 1.0$\cite{SID2} & 1 & 1.499 & 10.188 \\
& SiD$^2$A, $\alpha = 1.2$ \cite{SID2}& 1 & 1.519 & 10.252 \\
& \textbf{SiD$^2$A+ EL (ours), $\alpha = 1.2$} & 1 & \textbf{1.475} & 10.23 \\
\bottomrule
\end{tabular}%
}
\end{minipage}%
\hfill
\begin{minipage}[t]{0.48\textwidth}
\centering
\caption{Analogous to Table 2 for CIFAR-10 (conditional). ``Direct generation'' and ``Distillation'' methods presented in the table requires one single NFE, and the teacher requires 35 NFE.}
\label{tab:cifar10_cond}
\footnotesize
\adjustbox{max width=\textwidth}{%
\begin{tabular}{llc}
\toprule
Family & Model & FID ($\downarrow$) \\
\midrule
Teacher & VP-EDM \cite{karrasElucidatingDesignSpace2022}& 1.79 \\
\midrule
\multirow{4}{*}{\begin{tabular}[c]{@{}l@{}}Direct\\generation\end{tabular}} 
& BigGAN\cite{bigGAN} & 14.73 \\
& StyleGAN2+ADA\cite{StyleGAN2+ADA+Tune}& 3.49 \\
& StyleGAN2+ADA+Tune\cite{StyleGAN2+ADA+Tune}& 2.42 \\
\midrule
\multirow{15}{*}{Distillation}
& GET-Base\cite{GET-Base} & 6.25 \\
& Diff-Instruct\cite{luoDiffInstructUniversalApproach2023}& 4.19 \\
& StyleGAN2+ADA+Tune+DI\cite{luoDiffInstructUniversalApproach2023}
& 2.27 \\
& DMD\cite{yinOnestepDiffusionDistribution2024} & 2.66 \\
& DMD (\textit{w.o. $\mathcal{L}_{adv}$})\cite{yinOnestepDiffusionDistribution2024} & 3.82 \\
& DMD (\textit{w.o. reg.}) \cite{yinOnestepDiffusionDistribution2024} & 5.58 \\
& CTM \cite{CTM} & 1.73 \\
& GDD-I\cite{GDD} & 1.44 \\
& SiD, $\alpha = 1.0$ \cite{SID}& 1.932 \\
& SiD, $\alpha = 1.2$\cite{SID} & 1.710\\
& SiDA, $\alpha = 1.0$\cite{SID2} & 1.436 \\
& SiD$^2$A, $\alpha = 1.0$\cite{SID2} & 1.403 \\
& \textbf{SiD$^2$A+ EL (ours), $\alpha = 1.0$} & 1.395 \\
& SiD$^2$A, $\alpha = 1.2$ \cite{SID2}& 1.396 \\
& \textbf{SiD$^2$A+ EL (ours), $\alpha = 1.2$} & \textbf{1.38} \\
\bottomrule
\end{tabular}%
}
\end{minipage}
\end{table*}

\begin{table*}[h]
\centering
\caption{Comparison of training efficiency and generation quality on FFHQ 64×64. The best resource-efficient one-step generator is highlighted with bold.}
\label{FFHQ_table}
\resizebox{0.97\textwidth}{!}{%
\begin{tabular}{llcccccc}
\toprule
Family & Model & NFE & FID ($\downarrow$) & Batch Size & Iterations & Iterated k-images & Device \\
\midrule
Teacher & VP-EDM\cite{karrasElucidatingDesignSpace2022} & 79 & 2.39 & 256 & 781K & 200000 & V100 * 8 \\
\midrule
\multirow{2}{*}{Diffusion} & VP-EDM\cite{karrasElucidatingDesignSpace2022} & 50 & 2.60 & 256 & 781K & 200000 & V100 * 8 \\
& Patch-Diffusion\cite{patch-Diffusion} & 50 & 3.11 & 512 & - & - & V100 * 16 \\
\midrule
\multirow{6}{*}{Distillation} & BOOT\cite{gu2023boot} & 1 & 9.00 & 128 & 500K & 64000 & A100 * 8 \\

& SiD, $\alpha = 1.2$ \cite{SID}& 1 & 1.550 & 512 & 977K & 500000 & A100 * 16 \\
& SiDA, $\alpha = 1.0$\cite{SID2}& 1 & 1.134 & 512 & 391K & 200000 & H100 * 8 \\
& SiD$^2$A, $\alpha = 1.0$ \cite{SID2}& 1 & \textbf{1.040} & 512 & 332K & 170000 & H100 * 8 \\
& SiD$^2$A, $\alpha = 1.2$\cite{SID2} & 1 & 1.109 & 512 & 156K & 80000 & H100 * 8 \\
& \textbf{SiD$^2$A+EL(ours), $\alpha = 1.0$} & 1 & 1.060 & \textbf{64} & \textbf{53K} & \textbf{26000} & 4090-24G * 2 \\

\bottomrule
\end{tabular}
}
\end{table*}

\begin{table}[htbp]
\centering
\caption{FID scores on ImageNet 512$\times$512 (XS model, 125M parameters) $^\dagger$ means method we reproduced.}
\label{tab:fid_xs}
\begin{tabular}{l c c c c}
\toprule
Method & CFG & NFE & FID ($\downarrow$) & Batch Size \\
\midrule
EDM2 & N & 63 & 3.53 & -- \\
EDM2 & Y & 63$\times$2 & 2.91 & -- \\
SiD & N & 1 & $3.353 $ & 2048 \\
SiDA & N & 1 & $2.228 $ & 2048 \\
SiD$^2$A & N & 1 & $2.156 $ & 2048 \\
SiD$^2$A$^\dagger$ & N & 1 & $2.191 $ & 2048 \\
\textbf{SiD$^2$A+EL(ours)} & N & 1 & $\textbf{2.132} $ & 2048 \\
SiD$^2$A$^\dagger$ & N & 1 & $2.684 $ & 16 \\
\textbf{SiD$^2$A+EL(ours)} & N & 1 & $2.371 $ & \textbf{16} \\
\bottomrule
\end{tabular}
\end{table}

\begin{figure}[htbp]
    \centering
    \begin{subfigure}{0.45\textwidth}
        \centering
        \includegraphics[width=\textwidth]{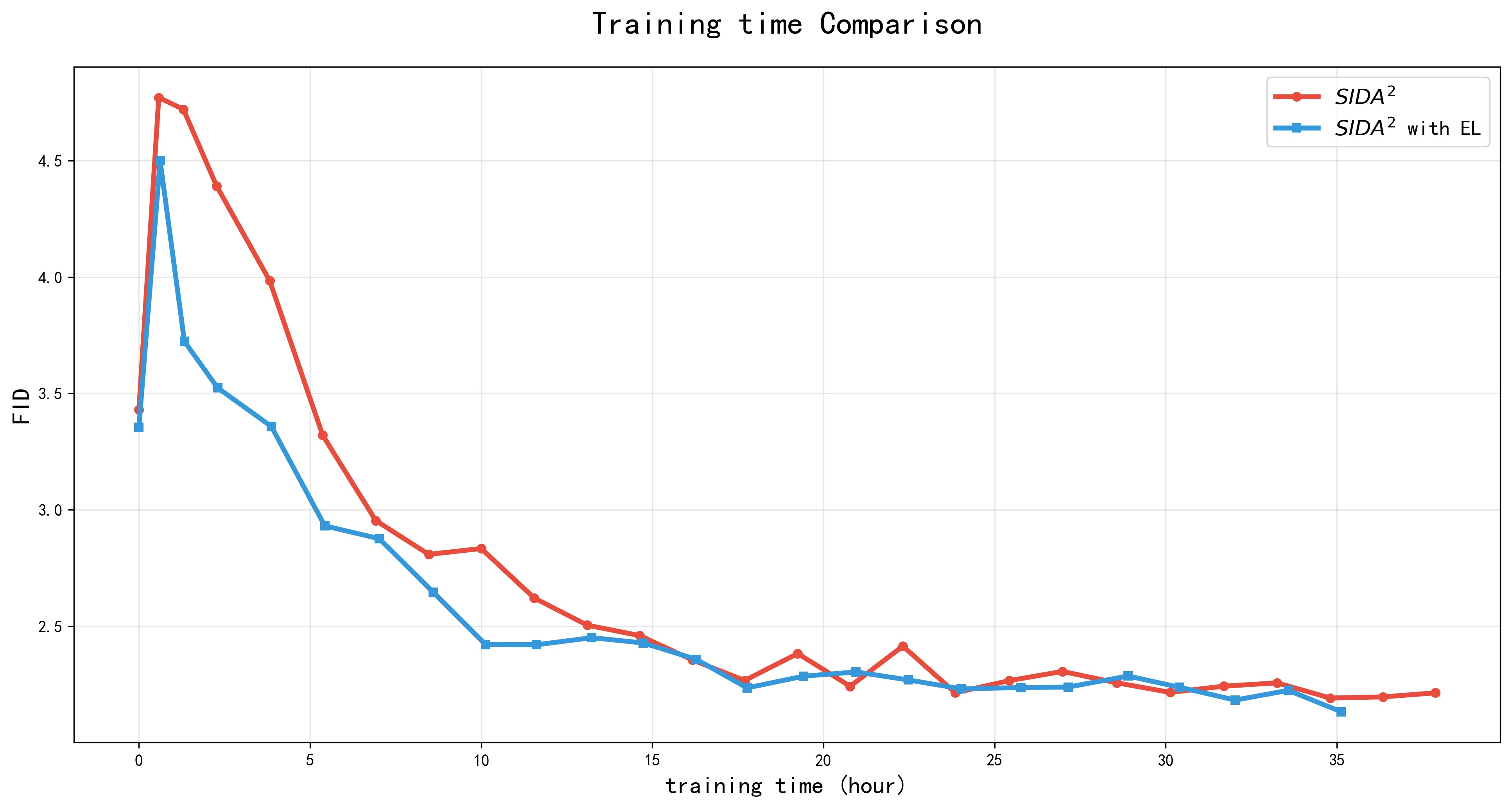}
        \caption{Batch size = 2048}
        \label{subfig:batch2048}
    \end{subfigure}
    \hfill
    \begin{subfigure}{0.45\textwidth}
        \centering
        \includegraphics[width=\textwidth]{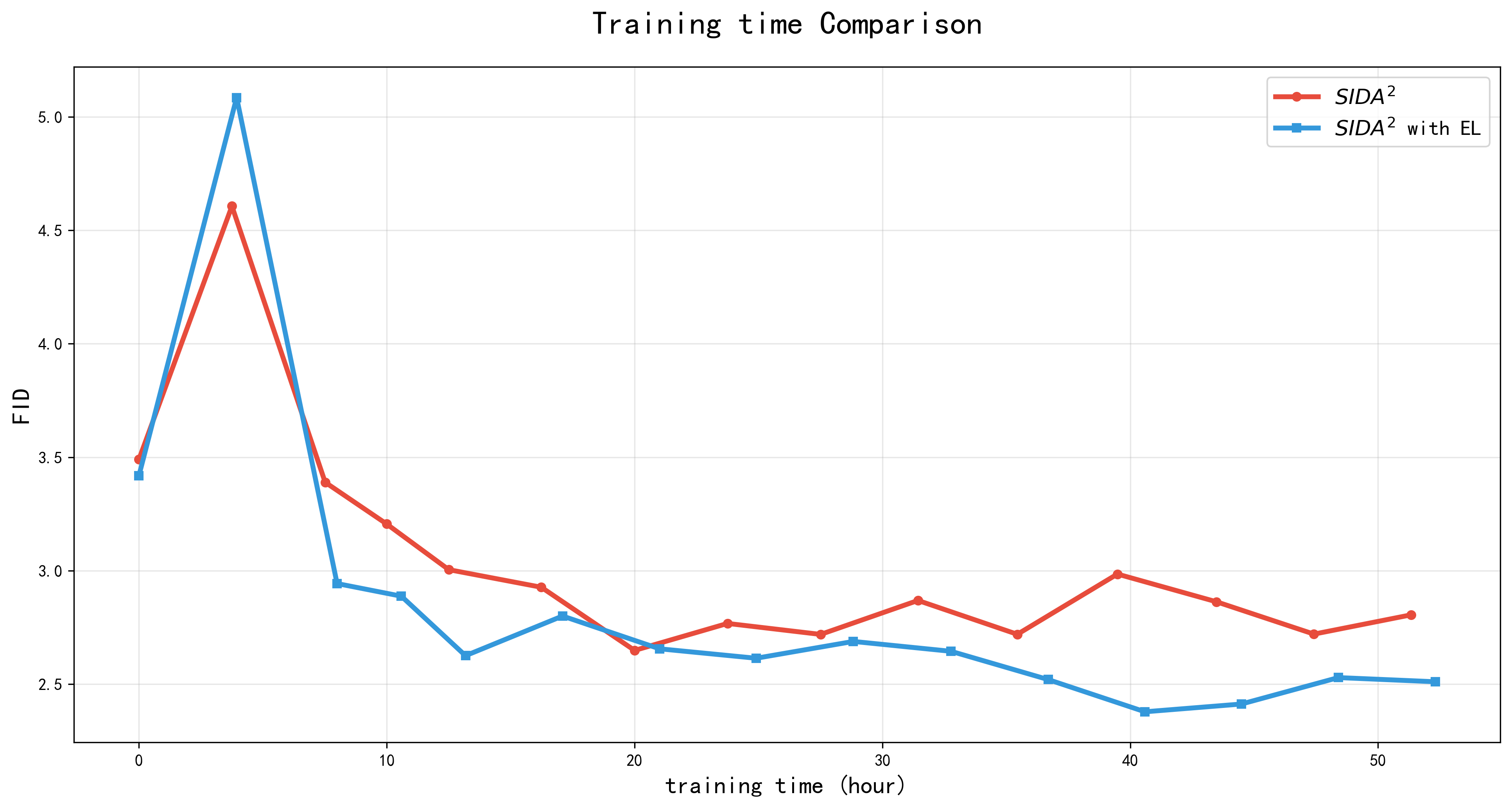}
        \caption{Batch size = 16}
        \label{subfig:batch16}
    \end{subfigure}
    \caption{SiD$^2$A training time comparison on ImageNet 512$\times$512.}
    \label{fig:fid_comparison}
\end{figure}

\section{Experiments}
We conduct comprehensive experiments to evaluate the effectiveness and efficiency of our proposed Embedding Loss (EL) for diffusion distillation. Initially, we demonstrate that distillation frameworks equipped with EL can rapidly achieve high-fidelity image generation while significantly reducing training time. Subsequently, we perform ablation studies to investigate the impact of the key parameters of EL, as well as the roles of other important hyperparameters. Finally, we systematically compare the performance of our method with existing distribution-matching and trajectory-preserving distillation approaches on standard benchmark datasets. Results consistently confirm that our EL not only stabilizes and accelerates training, but also generalizes well to various distillation frameworks, including distribution-matching distillation and trajectory-preserving distillation.

\subsection{Experimental Settings}
\subsection*{Datasets}
We assess EL's effectiveness across four standard benchmarks from EDM \cite{karrasElucidatingDesignSpace2022}: CIFAR-10 32×32 (cond/uncond) \cite{cifar10}, ImageNet 64×64,512×512 \cite{imagenet}, FFHQ 64×64 \cite{FFHQ}, and AFHQ-v2 64×64 \cite{AFHQ}.

\subsection*{Distillation Setup}
In this experiment, we apply DMD \cite{yinOnestepDiffusionDistribution2024}, DI \cite{luoDiffInstructUniversalApproach2023}, and SiD\textsuperscript{2}A \cite{SID2} with EL to distill pre-trained EDM \cite{karrasElucidatingDesignSpace2022} diffusion models into one-step generator models. Following the experimental setup of SiD\textsuperscript{2}A \cite{SID2}, we perform model distillation on the four datasets mention above. Detailed configurations are provided in Appendix \ref{Training and Evaluation Details and Additional Results}. We utilize the high-quality open-source codebase of SiD \cite{SID}.

\subsection*{Implementation Details}
We implement enhanced SiD\textsuperscript{2}A \cite{SID2}, DMD \cite{yinOnestepDiffusionDistribution2024} and DI \cite{luoDiffInstructUniversalApproach2023} with EL based on the EDM \cite{karrasElucidatingDesignSpace2022} codebase, and initialize both the generator $G_\theta$ and its score estimation network $f_\psi$ by copying the architecture and parameters of the pretrained score network $f_\phi$ from EDM\cite{karrasElucidatingDesignSpace2022}. Other implementation details are provided in Appendix~\ref{Training and Evaluation Details and Additional Results}.

\subsection*{Ablation Study}
\label{sec:ablation}
\subsubsection*{Comparison with Alternative Auxiliary Losses.}
To provide a fair comparison with regression loss and adversarial loss, we replace the regression loss component with EL while keeping all other experimental settings identical. As shown in Appendix \ref{Training and Evaluation Details and Additional Results}, EL consistently outperforms alternatives in training efficiency, while avoiding the computational overhead of dataset pre-generation and the training instability of adversarial optimization.

\subsubsection*{Ablation Study on Embedding Diversity}

\begin{table}[h]
\centering
\caption{Ablation study on architecture and initialization diversity}
\begin{tabular}{lcc}
\toprule
Architecture & Initialization & FID$\downarrow$ \\
\midrule
CNN Only & 1 Init & 4.45 \\
CNN Only & 4 Init & 4.30 \\
4 Different Arch & 1 Init & 4.1 \\
4 Different Arch & 4 Init & \textbf{3.95} \\
\bottomrule
\end{tabular}
\end{table}

To validate the effectiveness of embedding diversity in EL, we conduct ablation experiments on unconditional CIFAR-10 generation using DI \cite{luoDiffInstructUniversalApproach2023} as the baseline. Results show that both architecture diversity and initialization diversity contribute to generation quality. Using 4 diverse architectures with 4 random initializations achieves the best FID of 3.95, representing approximately 10\% improvement over the CNN-only baseline (4.45). This confirms that diverse architectures capture complementary features across different inductive biases, while varied initializations expand feature space coverage for robust distribution matching.

\subsection*{Applying EL to Alternative Frameworks}
To validate the generality of EL, we apply it to Consistency Distillation \cite{songConsistencyModels2023}, see Table \ref{CM:cifar10_comparison}. Results show that EL consistently improves both training efficiency and final performance.When added to the CD framework as an auxiliary loss, EL reduces training time by approximately 80\% to reach comparable convergence under the same experimental setup. In the 4-step generation setting, the distilled model with EL matches the teacher model's performance despite using significantly fewer sampling steps.These results indicate that EL provides effective regularization across different distillation methods. By aligning the student's output distribution with the data distribution, EL complements existing distillation methods' objectives and accelerates convergence.

\subsection{Benchmark Performance}
Our comprehensive evaluation compares the proposed method against most existing distribution-matching approaches and other leading deep generative models. All experimental results demonstrate that methods augmented with our Embedding Loss (EL) consistently outperform their EL-free counterparts in both final performance metrics and convergence speed.
In particular, models incorporating EL achieve statistically significant improvements in standard quantitative metrics, including Fréchet Inception Distance (FID) \cite{FID}, while requiring substantially fewer training iterations to reach convergence. To further prove the efficiency of EL, we show the comparison figure of training time and FID in \ref{subfig:batch2048}, from which it can be seen that EL can improve minimum performance and achieve faster convergence even when using a large batch size (2048) and improve performance more and achieve faster convergence when using a small batch (16). Random images generated by distribution matching distillation framework with EL in a single step are displayed in \cref{Unconditional CIFAR-10 photo,FFHQ 64 × 64 photo}.


\section{Conclusion}
We present Embedding Loss (EL), an innovative supplementary loss function that enables efficient distillation of pretrained diffusion models into high-quality few-step generators. By computing Maximum Mean Discrepancy in a diversified embedding space, EL achieves comprehensive distribution matching between generated samples and real data while maintaining training stability. Experimental results demonstrate EL's capability to significantly reduce Fréchet Inception Distance with remarkable efficiency, outperforming established distillation approaches across various configurations. This superiority extends to different distillation paradigms, including both distribution-matching and trajectory-preserving frameworks, and remains consistent regardless of the number of sampling steps or the need for additional regularization.

\bibliographystyle{unsrt}  
\bibliography{references}  

\newpage
\appendix
\onecolumn
\part*{Appendix for EL}
\section{Convergence Speed Comparison}

This section presents the convergence behavior comparison between models with and without the proposed Embedding Loss (EL). We evaluate the convergence speed on two different datasets and architectures to demonstrate the effectiveness of our approach.

The left figure shows the FID score evolution during training for the DI model on CIFAR-10, while the right figure displays results for SiD²A on FFHQ 64×64. As can be observed, incorporating the Embedding Loss consistently accelerates convergence and leads to better final performance across different model architectures and datasets. The EL-enhanced models reach lower FID scores faster and maintain more stable training dynamics, particularly in the early training stages.

\begin{figure}[htbp]
    \centering
    \begin{minipage}{0.45\textwidth}
        \centering
        \includegraphics[width=\textwidth]{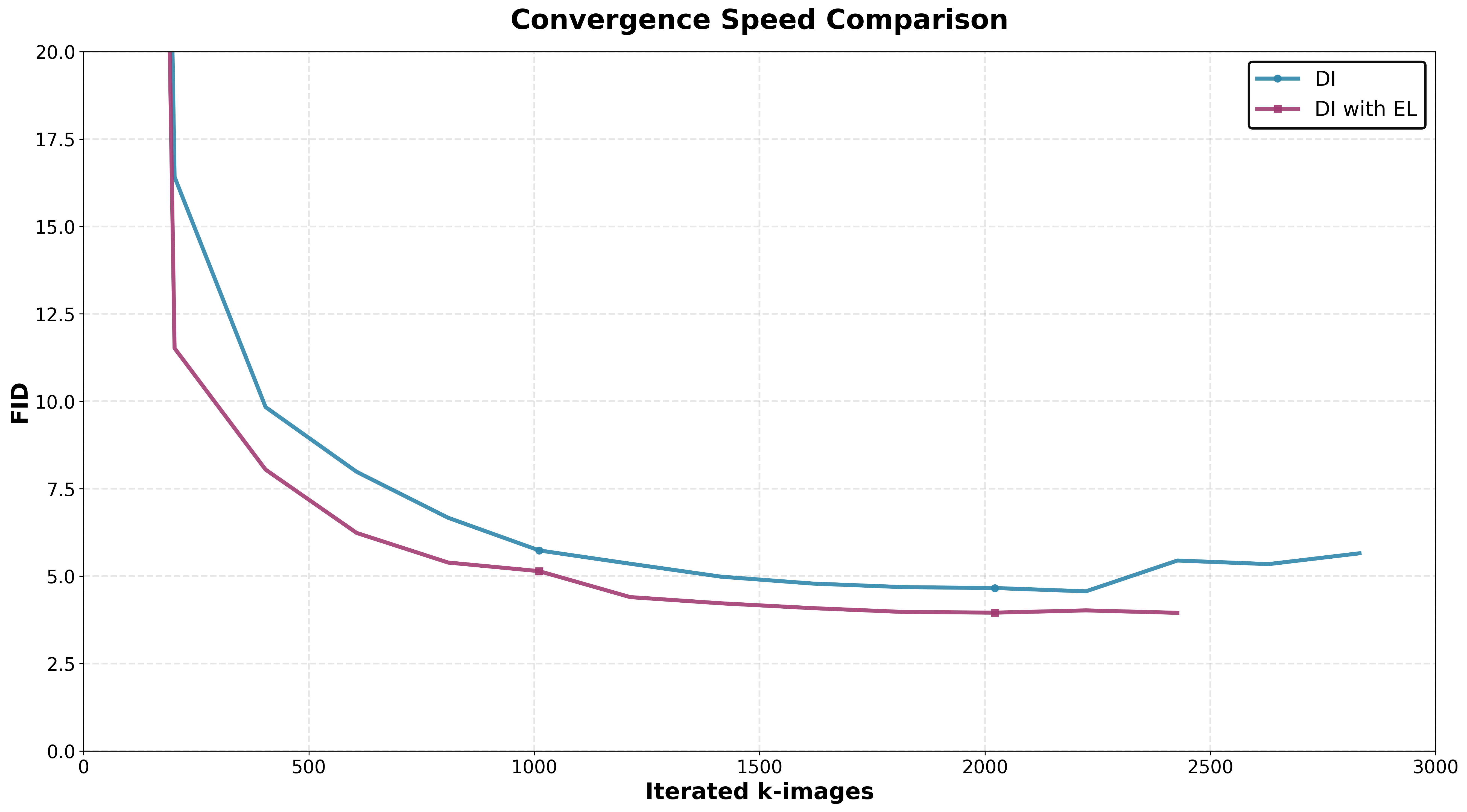}
        \caption{DI Convergence Speed Comparison on CIFAR-10}
    \end{minipage}
    \hfill
    \begin{minipage}{0.45\textwidth}
        \centering
        \includegraphics[width=\textwidth]{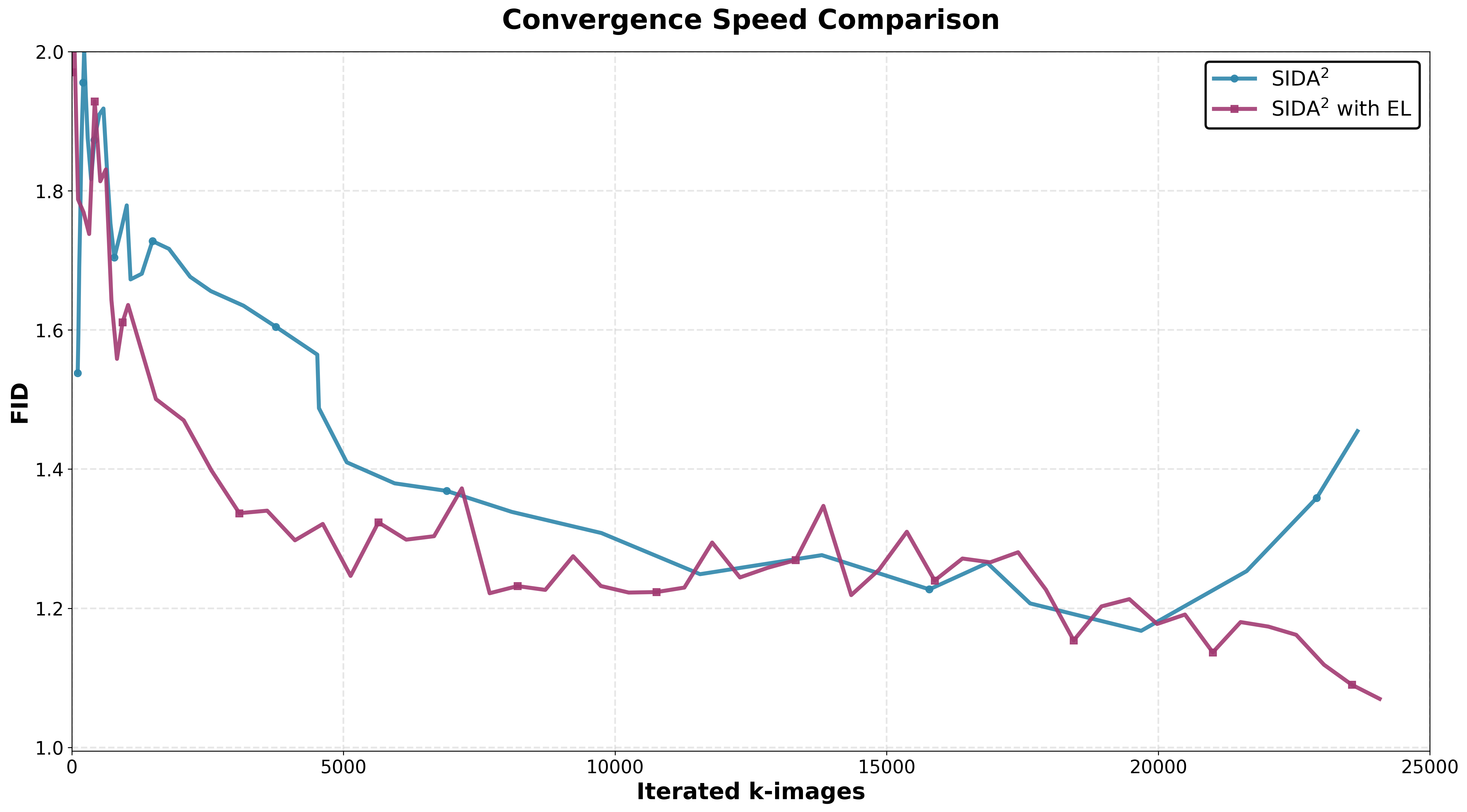}
        \caption{SiD$^2$A Convergence Speed Comparison on FFHQ 64 × 64}
    \end{minipage}
\end{figure}

\section{Training and Evaluation Details and Additional Results}
\label{Training and Evaluation Details and Additional Results}

\begin{table*}[h]
\centering
\caption{Hyperparameter configurations and performance comparison of CD-based\cite{songConsistencyModels2023} methods on CIFAR-10. The best resource-efficient part is highlighted with \textbf{bold}.}
\label{CM:cifar10_comparison}

\resizebox{0.95\textwidth}{!}{%
\begin{tabular}{lcccc}
\toprule
\textbf{Hyperparameter} & \textbf{Teacher (VP-EDM)}\cite{karrasElucidatingDesignSpace2022} & \textbf{CIFAR-10-Uncond-CD} & \textbf{CIFAR-10-Uncond-CD (smaller batch size)} & \textbf{CIFAR-10-Uncond-CD-with EL (ours)} \\
\midrule
Learning rate & - & 4e-4 & 4e-4 & 4e-4 \\
Batch size & - & 512 & \textbf{64} & \textbf{64} \\
$\mu$ & - & 0 & 0.95 & 0.95 \\
$N$ & - & 18 & 18 & 18 \\
EMA decay rate & - & 0.9999 & No & No \\
Training iterations & - & 800k & 200k & \textbf{120K} \\
Mixed-Precision (FP16) & - & No & No & No \\
Dropout probability & - & 0.0 & 0.0 & 0.0 \\
Number of GPUs & - & 8$\times$A100-40G & \textbf{1$\times$4090-24G} & \textbf{1$\times$4090-24G} \\
\midrule
NFE & 35 & 1 / 2 & 1 / 4 & 1 / 4 \\
FID & 2.04 & 3.55 / 2.93 & 4.8 / 3.3 & 3.5 / \textbf{2.04} \\
\bottomrule
\end{tabular}%
}
\end{table*}

\begin{table}[htbp]
\centering
\begin{minipage}{0.48\textwidth}
\centering
\caption{Hyperparameter settings for CIFAR-10 experiments}
\label{tab:hyperparameters_cifar10}
\resizebox{\textwidth}{!}{
\begin{tabular}{lccc}
\toprule
\textbf{Hyperparameter} & \textbf{SiD$^2$A-with-EL} & \textbf{SiD$^2$A-with-EL} & \textbf{DI-with-EL} \\
& \textbf{(Uncond)} & \textbf{(Cond)} & \textbf{(Uncond)} \\
\midrule
Learning rate & 1e-5 & 1e-5 & 1e-5 \\
Batch size & 64 & 64 & 128 \\
Gradient accumulation round & 4 & 4 & 1 \\
$\sigma(t^*)$ & 2.5 & 2.5 & 1 \\
Adam $\beta_0$ & 0.0 & 0.0 & 0.0 \\
Adam $\beta_1$ & 0.999 & 0.999 & 0.999 \\
fp16 & False & False & False \\
augment, dropout, cres & \multicolumn{3}{c}{Same as in EDM and SiD} \\
$\lambda_{\text{emd}}$ & 10 & 10 & 10 \\
$d$ & 64 & 64 & 64 \\
GPUs & $2 \times$ 4090-24G & $2 \times$ 4090-24G & $1 \times$ 4090-24G \\
num\_networks\_per\_type & 1 & 1 & 1 \\
\bottomrule
\end{tabular}
}
\end{minipage}
\hfill
\begin{minipage}{0.48\textwidth}
\centering
\caption{Hyperparameter settings for 64 X 64 experiments}
\label{tab:hyperparameters_64x64}
\resizebox{\textwidth}{!}{
\begin{tabular}{lcc}
\toprule
\textbf{Hyperparameter} & \textbf{SiD$^2$A-with-EL} & \textbf{SiD$^2$A-with-EL} \\
& \textbf{(FFHQ)} & \textbf{(AFHQ-V2)} \\
\midrule
Learning rate & 1e-5 & 5e-6 \\
Batch size & 64 & 64 \\
Gradient accumulation round & 8 & 8 \\
$\sigma(t^*)$ & 2.5 & 2.5 \\
Adam $\beta_0$ & 0.0 & 0.0 \\
Adam $\beta_1$ & 0.999 & 0.999 \\
fp16 & True & True \\
augment, dropout, cres & \multicolumn{2}{c}{Same as in EDM and SiD} \\
$\lambda_{\text{emd}}$ & 10 & 10 \\
$d$ & 64 & 64 \\
GPUs & $2 \times$ 4090-24G & $2 \times$ 4090-24G \\
num\_networks\_per\_type & 1 & 1 \\
\bottomrule
\end{tabular}
}
\end{minipage}
\end{table}

\begin{table}[h]
\centering
\caption{Quantitative comparison of generative models on ImageNet-64$\times$64. Best results in each category are highlighted in bold.}
\label{tab:imagenet_comparison}
\begin{tabular}{lcccccc}
\toprule
Method & \# Fwd & FID & Batch & Iterations & Iterated & Training \\
 & Pass ($\downarrow$) & ($\downarrow$) & Size &  &  M-images & Hardware \\
\midrule
BigGAN-deep \cite{bigGAN} & 1 & 4.06 & 2048 & 200K & 409.6 & 8$\times$TPUv3 \\
ADM \cite{beatsGAN} & 250 & 2.07 & 768 & 2000K & 1536.0 & 8$\times$V100 \\
RIN \cite{RIN} & 1000 & 1.23 & 1024 & 300K & 307.2 & 32$\times$TPUv3 \\
StyleGAN-XL \cite{stylegan_XL} & 1 & 1.52 & -- & -- & -- & -- \\
\midrule
Progress. Distill. \cite{salimansProgressiveDistillationFast2022} & 1 & 15.39 & 2048 & 550K & 1126.4 & 8$\times$TPUv4 \\
DFNO \cite{DFNO} & 1 & 7.83 & 2048 & 400K & 819.2 & -- \\
BOOT \cite{gu2023boot} & 1 & 16.30 & 1024 & 300K & 307.2 & 8$\times$A100 \\
TRACT \cite{TRACT} & 1 & 7.43 & 512 & 125K & 64.0 & 8$\times$A100 \\
Meng et al. \cite{meng} & 1 & 7.54 & 512 & -- & -- & -- \\
Diff-Instruct \cite{luoDiffInstructUniversalApproach2023} & 1 & 5.57 & 96 & -- & -- & 8$\times$V100 \\
Consistency Model \cite{songConsistencyModels2023} & 1 & 6.20 & 2048 & 600K & 1228.8 & 64$\times$A100 \\
iCT-deep \cite{song2023improvedCM} & 1 & 3.25 & 4096 & 800K & 3276.8 & N$\times$A100 \\
CTM \cite{Consistency_trajectory_models} & 1 & 1.92 & 2048 & 30K & 61.4 & 8$\times$A100 \\
DMD \cite{yinOnestepDiffusionDistribution2024} (Reg loss) & 1 & 2.62 & 336 & 350K & 117.6 & 7$\times$A100 \\
\textbf{DMD+EL (ours) (Emd loss)} & 1 & 2.25 & \textbf{16} & \textbf{200K} & \textbf{3.2} & \textbf{1$\times$RTX4090} \\
DMD2 \cite{yinImprovedDistributionMatching2024} (Adv loss) & 1 & \textbf{1.51} & 280 & 200K & 56.0 & 7$\times$A100 \\
\midrule
EDM (Teacher, ODE) \cite{karrasElucidatingDesignSpace2022} & 511 & 2.32 & 4096 & 600K & 2457.6 & 32$\times$A100 \\
EDM (Teacher, SDE) \cite{karrasElucidatingDesignSpace2022} & 511 & \textbf{1.36} & 4096 & 600K & 2457.6 & 32$\times$A100 \\
\bottomrule
\end{tabular}
\end{table}

\begin{table}[htbp]
\centering
\caption{Performance and Efficiency Comparison of Diffusion Distillation Methods on AFHQ-V2 64$\times$64. Best results in each category are highlighted in bold.}
\label{tab:AFHQ_comparison}
\begin{tabular}{llcccccc}
\toprule
Family & Model & NFE & FID & Batch & Iterations & Iterated & Device \\
 &  &  & ($\downarrow$) & Size &  & k-images & \\
\midrule
Teacher & VP-EDM \cite{karrasElucidatingDesignSpace2022} & 79 & 1.96 & 256 & 781K & 200K & 8$\times$A100 \\
\midrule
\multirow{6}{*}{Distillation} & SiD\cite{SID}, $\alpha = 1.0$ & \textbf{1} & 1.628 & 256 & 1445K & 370K & 16$\times$A100 \\
 & SiD\cite{SID}, $\alpha = 1.2$ & \textbf{1} & 1.711 & 256 & 1172K & 300K & 16$\times$A100 \\
 & SiDA\cite{SID2}, $\alpha = 1.0$  & \textbf{1} & 1.345 & 512 & 254K & 130K & 8$\times$A100 \\
 & SiD$^2$A\cite{SID2}, $\alpha = 1.0$  & \textbf{1} & 1.276 & 512 & 332K & 170K & 8$\times$A100 \\
 & SiD$^2$A\cite{SID2}, $\alpha = 1.2$  & \textbf{1} & 1.366 & 512 & 332K & 170K & 8$\times$A100 \\
 & \textbf{SiD$^2$A+EL (ours)}, $\alpha = 1.0$  & \textbf{1} & 1.30 & \textbf{64} & \textbf{78K} & \textbf{40K} & \textbf{2$\times$RTX4090} \\
 & \textbf{SiD$^2$A+EL (ours), $\alpha = 1.0$ (longer training)} & \textbf{1} & \textbf{1.26} & \textbf{64} & 320K & 160K & \textbf{2$\times$RTX4090} \\
\bottomrule
\end{tabular}
\end{table}
\FloatBarrier

\section{Theoretical Proofs and Derivations}
\label{app:theory}

\subsection{Notation}
\label{app:notation}

\begin{itemize}
\item $\phi$: teacher network parameters
\item $\theta$: student network parameters  
\item $s_{\phi}, s_{\theta}$: teacher and student score functions
\item $G_{\theta}$: student generator
\item $x_t$: noisy sample at timestep $t$
\item $z \sim p_z$: initial random noise from $p_z$
\item $p_t(x_t)$: marginal distribution of $x_t$ at step $t$
\item $q_t(x_t | x_0)$: conditional distribution given clean sample $x_0$
\item $\nabla_{x_t}$: gradient operator w.r.t. $x_t$
\item $\mathbb{E}_{x \sim p}[\cdot]$: expectation over distribution $p$
\end{itemize}

\subsection{Proof of Proposition 1}
\label{app:prop1_proof}

\textbf{Proposition 1 (Gap Propagation in Distribution Matching):}
Consider distribution matching distillation with the following objective.  
We define the score gap as  
\begin{equation}
\Delta(x_t, t) := s_{\theta}(x_t, t) - s_{\phi}(x_t,t),
\end{equation}  
where $\Delta(x_t, t)$ denotes the discrepancy between the student score function $s_{\theta}(x_t, t)$ and the teacher score function $s_{\phi}(x_t,t)$.
\begin{equation}
\mathcal{L}_{\text{DM}}(\theta) = \int_{t=0}^{T} w(t) \mathbb{E}_{z, x_0, x_t} \left\| \Delta(x_t, t) \right\|^2  dt,
\end{equation}
where $x_0 = G_\theta(z)$ with $z \sim p_z$, $x_t \mid x_0 \sim q_t(x_t \mid x_0)$ , and  $\left\| \Delta(x_t, t) \right\|^2$ denotes the squared Euclidean norm of the score gap. 

Let $\theta^*$ denote a local minimum of $\mathcal{L}_{\text{DM}}$.Under Assumption 1, the induced student distribution satisfies:
\begin{equation}
D_{KL}(p_{\text{data}} \| p_{\theta^*}) \leq C_1 \epsilon_{\text{teacher}}^2 + C_2 \epsilon_{\text{opt}}
\end{equation}
where $\epsilon_{\text{teacher}}^2 = \mathbb{E}_{t,x_t}[\|\Delta(x_t, t)\|^2]$ is the teacher's score approximation error, $\epsilon_{\text{opt}}$ is the student's optimization error, and $C_1, C_2$ are constants depending on $w(t)$ and the Lipschitz constant of the score functions.

\begin{proof}
    This proof derives an upper bound on the KL divergence between the student and data distributions via score matching theory, linking the distribution matching distillation loss to the KL divergence. The core idea is to start from the score gap, combine the teacher model error and student optimization error, accumulate errors through Fisher divergence and time integration, and finally separate error sources to obtain a linear upper bound. We elaborate on the detailed derivation below.  
	
	First, we review the necessary background and notation. The teacher model has parameters $\phi$ with score function $s_{\phi}(x_t, t) \approx \nabla_{x_t} \log p_t(x_t)$; the student model has parameters $\theta$ with score function $s_{\theta}(x_t, t)$. Define the score gap $\Delta(x_t, t) = s_{\theta}(x_t, t) - s_{\phi}(x_t, t)$, and the distillation objective (distribution matching loss) as:  
	\begin{equation}
		\mathcal{L}_{\text{DM}}(\theta) = \int_{t=0}^{T} w(t) \mathbb{E}_{z, x_0, x_t} \left[ \left\| \Delta(x_t, t) \right\|^2 \right] dt,
	\end{equation}  
	where $x_0 = G_{\theta}(z),\ z \sim p_z,\ x_t \mid x_0 \sim q_t(x_t \mid x_0)$.

	The gradient of $\mathcal{L}_{\text{DM}}$ with respect to $\theta$ is:

	Assumption 1 (well-trained teacher) states that the teacher's score approximates the true score with bounded error:  
	\begin{equation}
		\mathbb{E}_{x_t \sim p_{\text{data}}(x_t|t)} \left[ \left\| s_{\phi}(x_t, t) - \nabla_{x_t} \log p_t(x_t) \right\|^2 \right] \leq \epsilon^2(t),
	\end{equation}  
	i.e. the error is bounded by $\epsilon(t)$. Let $\theta^*$ be a local minimum of $\mathcal{L}_{\text{DM}}$. We aim to show $D_{\mathrm{KL}}(p_{\text{data}} \| p_{\theta^*}) \leq C_1 \epsilon_{\text{teacher}}^2 + C_2 \epsilon_{\text{opt}}$, where $\epsilon_{\text{teacher}}^2 = \mathbb{E}_{t,x_t}[\|\Delta(x_t,t)\|^2]$ (teacher's score approximation error) and $\epsilon_{\text{opt}}$ is the student's optimization error.  
	
	Next, we outline the proof strategy: Starting from the relationship between the distribution matching loss and the KL divergence, decompose the KL divergence into error contributions across different time steps (variational inference perspective), use the connection between score matching and KL divergence (Fisher divergence/score matching identity), and combine the teacher error $\epsilon_{\text{teacher}}$ with the student optimization error $\epsilon_{\text{opt}}$ to derive the upper bound.  
	
	The derivation begins with converting the score gap to a distribution gap. For smooth differentiable distributions $p, q$, the KL divergence is defined as $D_{\mathrm{KL}}(p \| q) = \mathbb{E}_{x \sim p} [\log p(x) - \log q(x)]$. In diffusion models, we consider the relationship between time-dependent marginal distributions $p_t(x_t)$ and $p_{\theta^*,t}(x_t)$. By applying a second-order Taylor expansion to the log-density functions, we obtain the Fisher divergence expression:
	\begin{equation}
		\frac{1}{2} \mathbb{E}_{x \sim p} \left\| \nabla_x \log p(x) - \nabla_x \log q(x) \right\|^2 = D_{\mathrm{KL}}(p \| q) + \text{constant term},
	\end{equation}  
	where the constant term depends only on the reference distribution $p$, which implies the score gap controls the KL divergence under stationary or specific conditions. In the forward diffusion process, the teacher marginal distribution is $p_t(x_t) = \int p_0(x_0) q_t(x_t \mid x_0) dx_0$, and the student marginal distribution is $p_{\theta^*,t}(x_t) = \int p_0'(x_0) q_t(x_t \mid x_0) dx_0$ ($p_0'$ is the student generator's distribution). The initial distribution KL divergence is decomposed into time-step-wise score matching errors via the chain rule.  
	
	We now proceed to the detailed derivation steps. Step 1 analyzes score error propagation: By Assumption 1, the teacher's score error satisfies $\left\| s_{\phi}(x_t, t) - \nabla_{x_t} \log p_t(x_t) \right\|^2 \leq \epsilon^2(t)$. Let $\epsilon_{\text{teacher}}^2 = \mathbb{E}_{t,x_t}[\|\Delta(x_t,t)\|^2]$ denote the score gap between the student and teacher. Then the student's score is $s_{\theta}(x_t, t) = s_{\phi}(x_t, t) + \Delta(x_t, t)$. By the triangle inequality and basic inequalities $\left\| a+b \right\|^2 \leq 2\left\| a \right\|^2 + 2\left\| b \right\|^2$, substituting gives:  
	\begin{equation}
		\begin{split}
			\left\| s_{\theta}(x_t, t) - \nabla_{x_t} \log p_t(x_t) \right\|^2 
			\leq 2\left\| s_{\phi} - \nabla_{x_t} \log p_t \right\|^2 + 2\left\| \Delta \right\|^2 &\leq 2\epsilon^2(t) + 2\epsilon_{\text{teacher}}^2.
		\end{split}
	\end{equation}  
	
	Step 2 links the score error to Fisher divergence. For any $t$, by score matching theory:  
	\begin{equation}
		\frac{1}{2} \mathbb{E}_{x_t \sim p_t} \left\| s_{\theta}(x_t, t) - \nabla_{x_t} \log p_t(x_t) \right\|^2 \geq D_{\mathrm{KL}}(p_t \| p_{\theta^*,t}) + \text{constant term}.
	\end{equation}  
	Ignoring the constant term, we obtain $D_{\mathrm{KL}}(p_t \| p_{\theta^*,t}) \leq C(t) (\epsilon^2(t) + \epsilon_{\text{teacher}}^2)$, where $C(t)$ depends on the Lipschitz constant of the score functions and the weight $w(t)$.  
	
	Step 3 accumulates the KL divergence from $p_t$ to the initial distribution. Using the Markov property of the diffusion process, total errors are accumulated to $t=0$ (initial data distribution): For discrete time, $D_{\mathrm{KL}}(p_{\text{data}} \| p_{\theta^*}) = \sum_{t=1}^T D_{\mathrm{KL}}(p_{t-1} \| p_{\theta^*,t-1})$; for continuous time, the integral form is:  
	\begin{equation}
		D_{\mathrm{KL}}(p_{\text{data}} \| p_{\theta^*}) \leq \int_0^T C(t) (\epsilon^2(t) + \epsilon_{\text{teacher}}^2) dt.
	\end{equation}  
	
	Step 4 separates the teacher error and optimization error. The teacher approximation error (difference between $s_{\phi}$ and the true score, $\epsilon^2(t)$) is absorbed into $C_1 \epsilon_{\text{teacher}}^2$ (since $\epsilon_{\text{teacher}}^2$ includes the student-teacher gap, and the teacher-true gap is controlled by Assumption 1). The optimization error arises because $\theta^*$ is not globally optimal, leaving a residual loss $\mathcal{L}_{\text{DM}}(\theta^*) = \epsilon_{\text{opt}}$. Thus:  
	\begin{equation}
		D_{\mathrm{KL}}(p_{\text{data}} \| p_{\theta^*}) \leq C_1 \epsilon_{\text{teacher}}^2 + C_2 \epsilon_{\text{opt}},
	\end{equation}  
	where $C_1 = \int_0^T C(t) dt$ (depending on $w(t)$ and Lipschitz constants) and $C_2$ captures optimization residual effects.  
	
	In summary, the KL divergence upper bound is:  
	\begin{equation}
		D_{\mathrm{KL}}(p_{\text{data}} \| p_{\theta^*}) \leq C_1 \epsilon_{\text{teacher}}^2 + C_2 \epsilon_{\text{opt}}
	\end{equation}
\end{proof}

\subsection{Detailed Variance Analysis}
\label{app:variance_analysis}

\textbf{Proposition 2 (Gradient Variance in DM):}
In distribution matching distillation with batch size $B$, the gradient estimate is:

\begin{equation}
\widehat{\text{Grad}}(\theta) = \frac{1}{B}\sum_{i=1}^B \frac{\partial}{\partial\theta} \mathcal{L}_{\text{DM}}(\theta) = \frac{1}{B}\sum_{i=1}^B \frac{\partial}{\partial\theta}\int_{t=0}^T w(t)\big[\{-\mathbf{d}'(y_t^{(i)})\}^T \{s_\theta(\boldsymbol{x}_t^{(i)},t) - \nabla_{x_t}\log q_t(\boldsymbol{x}_t^{(i)}|\boldsymbol{x}_0^{(i)})\}\big]dt
\end{equation}

where $t$ is the diffusion time variable ($0 \leq t \leq T$, $T$ total steps), $B$ the batch size, $w(t)$ a time-weighting function balancing diffusion stage contributions, $\mathbf{d}(\cdot)$ a distance function (commonly Euclidean) mapping targets to data space with $\mathbf{d}'(\cdot)$ its derivative (or Jacobian transpose), $s_\theta(\boldsymbol{x}_t^{(i)}, t)$ the score network (param. $\theta$) estimating $\nabla_{\boldsymbol{x}_t} \log q_t(\boldsymbol{x}_t)$, $q_t(\boldsymbol{x}_t^{(i)}|\boldsymbol{x}_0^{(i)})$ the conditional density of noisy sample $\boldsymbol{x}_t^{(i)}$ given clean $\boldsymbol{x}_0^{(i)}$ in forward diffusion, and $\boldsymbol{x}_t^{(i)} \sim q_t(\cdot|\boldsymbol{x}_0^{(i)})$ indicating sampling from $q_t$.

The variance satisfies:
\begin{equation}
\begin{split}
\text{Var}(\widehat{\text{Grad}}(\theta)) = \sigma^2_{\text{noise}} + \sigma^2_{\text{time}}
+ \sigma^2_{\text{diffusion}} + \mathcal{O}(1/B)
\end{split}
\end{equation}

where the variance decomposes into four distinct sources:
\begin{itemize}
\item $\sigma^2_{\text{noise}}$: variance from random noise $\boldsymbol{\epsilon} \sim \mathcal{N}(0, I)$ in the forward diffusion process $q_t(\boldsymbol{x}_t|\boldsymbol{x}_0)$, which introduces stochasticity independent of the generated sample;

\item $\sigma^2_{\text{time}}$: variance from random timestep sampling $t \sim \mathcal{U}[0,T]$, as different timesteps have different denoising difficulties and gradient magnitudes;

\item $\sigma^2_{\text{diffusion}}$: variance inherent to the teacher score network $s_\phi(\boldsymbol{x}_t, t)$ approximation error and the continuous-time integral discretization in practice;

\item $\mathcal{O}(1/B)$: standard Monte Carlo variance that decreases with batch size $B$, arising from finite-sample averaging over the latent distribution $p_z$.
\end{itemize}
Critically, the first three variance terms are \textit{independent of batch size}, explaining why distribution matching methods require large batches to overcome these irreducible noise sources.

\begin{proof}
    The variance decomposition of the gradient estimator involves three steps: application of the Law of Total Variance, splitting of time-step and diffusion variances, and analysis of variance from batch approximation.
	
	\subsubsection*{Step 1: Initial Decomposition via Law of Total Variance}
	The Law of Total Variance states that for any random variable $Y$ and conditioning variable $Z$, we have:
	\[
	\text{Var}(Y) = \mathbb{E}\left[ \text{Var}(Y \mid Z) \right] + \text{Var}\left( \mathbb{E}\left[ Y \mid Z \right] \right)
	\]
	
	Let $Y = \widehat{\text{Grad}}(\theta)$ and the conditioning variable $Z$ be the noise $z$. Then:
	\[
	\text{Var}(\widehat{\text{Grad}}(\theta)) = \underbrace{\mathbb{E}\left[ \text{Var}\left( \widehat{\text{Grad}}(\theta) \mid z \right) \right]}_{\sigma^2_{\text{noise}}} + \underbrace{\text{Var}\left( \mathbb{E}\left[ \widehat{\text{Grad}}(\theta) \mid z \right] \right)}_{\text{residual variance}}
	\]
	
	Here, $\sigma^2_{\text{noise}}$ is the expectation of the variance of the remaining randomness (time step, samples) given the noise $z$, i.e., the variance contribution from the noise itself.
	
	\subsubsection*{Step 2: Time-Step and Diffusion Decomposition of Residual Variance}
	Applying the Law of Total Variance again to the residual variance $\text{Var}\left( \mathbb{E}\left[ \widehat{\text{Grad}}(\theta) \mid z \right] \right)$, with the conditioning variable as the time step $t$:
	\[
	\text{Var}\left( \mathbb{E}\left[ \widehat{\text{Grad}}(\theta) \mid z \right] \right) = \mathbb{E}_t\left[ \text{Var}\left( \mathbb{E}\left[ \widehat{\text{Grad}}(\theta) \mid z, t \right] \mid t \right) \right] + \text{Var}_t\left( \mathbb{E}\left[ \widehat{\text{Grad}}(\theta) \mid z, t \right] \right)
	\]
	
	\begin{itemize}
		\item First term $\text{Var}_t\left( \mathbb{E}\left[ \widehat{\text{Grad}}(\theta) \mid z, t \right] \right)$: The variance of the expectation of the gradient estimator with respect to $t$, given $z$ and $t$, denoted as $\sigma^2_{\text{time}}$ (time-step variance).
		
		\item Second term $\mathbb{E}_t\left[ \text{Var}\left( \mathbb{E}\left[ \widehat{\text{Grad}}(\theta) \mid z, t \right] \mid t \right) \right]$: The conditional variance of the gradient estimator given $t$, after taking the expectation over $t$, arising from the randomness of the forward diffusion process, denoted as $\sigma^2_{\text{diffusion}}$ (diffusion variance).
	\end{itemize}
	
	\subsubsection*{Step 3: Variance from Mini-Batch Approximation ($\mathcal{O}(1/B)$ Term)}
	In actual training, the gradient is estimated via a mini-batch (batch size $B$): for $B$ independent samples, the gradient estimator is $\frac{1}{B} \sum_{i=1}^B \widehat{\text{Grad}}_i(\theta)$ ($\widehat{\text{Grad}}_i(\theta)$ is the gradient estimate for the $i$-th sample).
	
	By the variance property of sums of independent random variables:
	\[
	\text{Var}\left( \frac{1}{B} \sum_{i=1}^B \widehat{\text{Grad}}_i(\theta) \right) = \frac{1}{B^2} \sum_{i=1}^B \text{Var}(\widehat{\text{Grad}}_i(\theta)) = \frac{1}{B} \cdot \text{Var}(\widehat{\text{Grad}}(\theta))
	\]
	
	Thus, the variance introduced by the mini-batch approximation is $\mathcal{O}\left( \frac{1}{B} \right)$ (which tends to 0 as $B \to \infty$).
	
	\subsubsection*{Step 4: Combining All Variance Terms}
	Combining the above decomposition results, we obtain:
	\[
	\text{Var}(\widehat{\text{Grad}}(\theta)) = \sigma^2_{\text{noise}} + \sigma^2_{\text{time}} + \sigma^2_{\text{diffusion}} + \mathcal{O}\left( \frac{1}{B} \right)
	\]
	
	\subsubsection*{Guarantee of Variance Boundedness}
	The boundedness of each variance component $\sigma^2_{\text{noise}}, \sigma^2_{\text{time}}, \sigma^2_{\text{diffusion}}$ is guaranteed by the following two points:
	\begin{enumerate}
		\item Lipschitz continuity of the score function: If the score function $s_\theta(x_t)$ and the forward score $\nabla \log q_t(x_t \mid x_0)$ satisfy the Lipschitz condition (i.e., $\| s_\theta(x_t) - s_\theta(\tilde{x}_t) \| \leq L \| x_t - \tilde{x}_t \|$), the fluctuation of the gradient estimate will be bounded.
		
		\item Boundedness of the weight function: If the gradient estimate involves a weight function $w(t)$ (e.g., the time weight in VeB-SDE), the boundedness of $w(t)$ (e.g., $\| w(t) \| \leq W$) further controls the variance growth.
	\end{enumerate}
	
	In conclusion, the variance decomposition and the conclusion regarding the $\mathcal{O}(1/B)$ term hold.
\end{proof}

\textbf{Corollary 1 (Batch Size Dependency):}
Assume the data distribution has finite second moments: $\mathbb{E}_{x \sim p_{\text{data}}}[\|x\|^2] < \infty$. Then the distillation error in total variation distance is bounded by:
\begin{equation}
D_{TV}(p_{\theta^*}, p_{\text{data}}) \leq D_{TV}(p_{\theta^*}, \hat{p}_{\text{data}}) + D_{TV}(\hat{p}_{\text{data}}, p_{\text{data}}).
\end{equation}

\begin{proof}
    \subsubsection*{Step 1: Total Variation Distance and the Triangle Inequality}
The total variation distance $D_{TV}(P, Q)$ between two probability distributions $P$ and $Q$ (on a common measurable space $(\Omega, \mathcal{F})$) quantifies their dissimilarity. It is defined as:
\begin{itemize}
    \item For discrete distributions: $D_{TV}(P, Q) = \frac{1}{2} \sum_{x \in \Omega} |P(x) - Q(x)|,$
    \item For absolutely continuous distributions with densities $p, q$: $D_{TV}(P, Q) = \frac{1}{2} \int_{\Omega} |p(x) - q(x)| \, dx.$
\end{itemize}
The total variation distance satisfies the triangle inequality (since it is a metric on the space of probability measures): for any distributions $P, Q, R,$
$$D_{TV}(P, R) \leq D_{TV}(P, Q) + D_{TV}(Q, R).$$

\subsubsection*{Step 2: Apply the Triangle Inequality to the Distillation Error}
Let $P^{\theta^*}$ denote the model distribution (learned by the student model), $\hat{P}_{\text{data}}$ denote the empirical distribution constructed from a batch of $B$ data samples, and $P_{\text{data}}$ denote the true data distribution.
Applying the triangle inequality with $P = P^{\theta^*}$, $Q = \hat{P}_{\text{data}}$, and $R = P_{\text{data}}$, we get:
$$D_{TV}(P^{\theta^*}, P_{\text{data}}) \leq D_{TV}(P^{\theta^*}, \hat{P}_{\text{data}}) + D_{TV}(\hat{P}_{\text{data}}, P_{\text{data}}).$$
\end{proof}

\textbf{Corollary 2 (Slow Convergence):}
Under standard SGD, the convergence rate satisfies:
\begin{equation}
    \mathbb{E}[\mathcal{L}(\theta_T)] - \mathcal{L}(\theta^*) \sim O\left( \frac{1}{\sqrt{BT}} \right) 
\end{equation}
where $T$ is the number of iterations, and $B$ is batch size.

\begin{proof}
    
Consider the score matching loss of diffusion models:
\[
\mathcal{L}_{\text{DM}}(\theta) = \int_0^T w(t) \, \mathbb{E}_{z, x_0, x_t} \left\| s_\theta(x_t, t) - s^*(x_t, t) \right\|^2 dt,
\]
where \( s^*(x_t, t) \) is the true score function, and \( s_\theta(x_t, t) \) is the model-predicted score. In non-convex scenarios, optimized using mini-batch stochastic gradient descent with mini-batch size \( B \) and learning rate schedule \( \eta_k = \frac{\eta_0}{\sqrt{k + 1}} \) (sublinear decay, balancing convergence and stability), total number of diffusion time steps \( K \) (corresponding to discrete step size \( \Delta t \), satisfying \( T = K \cdot \Delta t \)).

The gradient estimate of the $k$-th batch (corresponding to time $t_k = k\Delta t$) is the batch average:  
\begin{equation*}
	\nabla \hat{\mathcal{L}}_{\text{DM},k}(\theta_t) = \frac{1}{B} \sum_{b=1}^B \nabla_\theta \left[ w(t) \| s_\theta(x_b^{(t)}) - s^*(x_b^{(t)}) \|^2 \right],
\end{equation*}
where $\{x_b^{(t)}\}_{b=1}^B$ are independent samples from the diffusion transition distribution $p(x_t | x_0)$.  

\subsubsection*{Expected Loss Update of Stochastic Gradient Descent}
Based on convexity tools, the expected loss of SGD iteration with batch size $B$ satisfies:
\begin{equation*}
	\mathbb{E}[L(\theta_{k+1})] - L(\theta^*) \leq \mathbb{E}[L(\theta_k)] - L(\theta^*) - \eta_k \mathbb{E}[\|\nabla L(\theta_k)\|^2] + \frac{\eta_k^2 L}{2} \cdot \text{Var}(\hat{\nabla} L(\theta_k)).
\end{equation*}

Substituting the upper bound of gradient variance $\text{Var}(\hat{\nabla} L(\theta_k)) = \frac{\sigma_{\text{full}}^2}{B}$, we get:
\begin{equation*}
	\mathbb{E}[L(\theta_{k+1})] - L(\theta^*) \leq \mathbb{E}[L(\theta_k)] - L(\theta^*) - \eta_k \mathbb{E}[\|\nabla L(\theta_k)\|^2] + \frac{\eta_k^2 L \sigma_{\text{full}}^2}{2B}.
\end{equation*}

\subsubsection*{Convergence Summation Over Multiple Iterations (Telescoping Sum)}
Summing over $k=0,1,\dots,K-1$ and rearranging the left-hand side using the telescoping sum method:
\begin{equation*}
	\sum_{k=0}^{K-1} \left( \mathbb{E}[L(\theta_{k+1})] - L(\theta^*) \right) \leq \sum_{k=0}^{K-1} \left( \mathbb{E}[L(\theta_k)] - L(\theta^*) - \eta_k \mathbb{E}[\|\nabla L(\theta_k)\|^2] + \frac{\eta_k^2 L \sigma_{\text{full}}^2}{2B} \right).
\end{equation*}

Noting that $\sum_{k=0}^{K-1} \left( \mathbb{E}[L(\theta_{k+1})] - L(\theta^*) \right) = \mathbb{E}[L(\theta_K)] - K L(\theta^*)$, and $\mathbb{E}[L(\theta_K)] \leq \sum_{k=0}^{K-1} \mathbb{E}[L(\theta_{k+1})] + L(\theta_0)$. Therefore:
\begin{equation*}
	\mathbb{E}[L(\theta_K)] - L(\theta^*) \leq \frac{\|\theta_0 - \theta^*\|^2}{2\eta K} + \frac{\sigma_{\text{full}}^2}{2\eta BK} + \sum_{k=0}^{K-1} \eta_k^2 \mathbb{E}[\|\nabla L(\theta_k) - \nabla L(\theta^*)\|^2].
\end{equation*}

\subsubsection*{Analysis of Each Error Term}
\begin{itemize}
	\item \textbf{Initial error term} $\frac{\|\theta_0 - \theta^*\|^2}{2\eta BK}$: The learning rate $\eta_k \propto \frac{1}{\sqrt{K}}$, so this term is $O\left( \frac{1}{\sqrt{K}} \right)$.
	
	\item \textbf{Gradient variance term} $\frac{\sigma_{\text{full}}^2}{2\eta BK}$: Since $\eta \propto \frac{1}{\sqrt{K}}$ and $\sigma_{\text{full}}^2$ is a constant independent of $B$ (determined by the data distribution and $w(t)$), this term is $O\left( \frac{1}{\sqrt{BK}} \right)$.
	
	\item \textbf{Gradient bias term} $\sum_{k=0}^{K-1} \eta_k^2 \mathbb{E}[\|\nabla L(\theta_k) - \nabla L(\theta^*)\|^2]$: In non-convex scenarios, gradient bias is mainly dominated by the multi-minima characteristics of local losses, with an order of $O\left( \frac{\log K}{\sqrt{K}} \right)$ (slower than the dominant order of the variance term).
\end{itemize}

\subsubsection*{Dominant Error Term}
The total diffusion time steps $T$ and iteration steps $K$ satisfy $T = K \cdot \Delta t$ ($\Delta t$ is a fixed discrete step size, so $K = O(T)$). Substituting $K$ with $T$ and ignoring lower-order terms (e.g. initial error, gradient bias), the dominant error term is the gradient variance term:
\begin{equation*}
	\mathbb{E}[L(\theta_T)] - L(\theta^*) \sim O\left( \frac{1}{\sqrt{BT}} \right).
\end{equation*}

\subsubsection*{Conclusions}
\begin{itemize}
	\item \textbf{Mechanism of Mini-batch Size $B$}: $B$ affects the convergence rate through the scaling of the variance term. The variance term changes from $O\left( \frac{1}{\sqrt{T}} \right)$ in full-sample SGD to $O\left( \frac{1}{\sqrt{BT}} \right)$. A larger $B$ (e.g., close to $N$) makes $\frac{1}{\sqrt{B}}$ smaller, leading to faster convergence.
	
	\item \textbf{Theoretical Significance of Convergence Order}: The rate $O\left( \frac{1}{\sqrt{BT}} \right)$ shows that training efficiency is jointly determined by $B$ and $T$. When $T$ is fixed, increasing $B$ can accelerate convergence; when $B$ is fixed, increasing $T$ (extending the diffusion training duration) can reduce error, but it is limited by computational resources and overfitting risks.
	
	\item \textbf{Rationality of Asymptotic Order}: Logarithmic factors or lower-order terms (e.g., gradient bias) are suppressed by the dominant term as $T \to \infty$ or $B \to \infty$, reflecting the universality of the mini-batch mechanism in non-convex diffusion models.
\end{itemize}

\subsubsection*{The Derivation Relies on the Following Core Assumptions}
\begin{itemize}
	\item The gradient of the score function $s_\theta$ satisfies Lipschitz continuity (ensuring smooth local losses);
	\item Independence of mini-batch sampling (unbiasedness of Monte Carlo gradients);
	\item Learning rates satisfy the Robbins–Monro conditions (unbiased estimation under adaptive decay).
\end{itemize}

In practice, the time weight scheduling of diffusion models (e.g., $w(t) \propto 1/\sigma_t^2$, where $\sigma_t$ is the forward noise standard deviation) and mini-batch strategies (e.g., dynamically adjusting $B$) can optimize the convergence constant by regulating $\sigma_{\text{full}}^2$ and computational resources, but cannot change the asymptotic order $O\left( \frac{1}{\sqrt{BT}} \right)$.

\end{proof}

\subsection{Analysis of Regression Loss}
\label{app:regression_analysis}

\textbf{Proposition 3 (Regression Loss Limitations):}
The gradient of regression loss satisfies:
\begin{equation}
\nabla_\theta \mathcal{L}_{\text{reg}}(\theta) = \mathbb{E}_{(z,y) \sim \mathcal{D}}\left[J_{G_\theta}(z)^T \cdot \nabla_x \ell(G_\theta(z), y)\right]
\end{equation}

This approach has three critical issues:

\textbf{1. Dependency on pre-generated dataset:} 
Requires constructing $\mathcal{D}$ offline using the teacher model with expensive deterministic sampling:
\begin{equation}
|\mathcal{D}| \gg B \quad \text{(typically } |\mathcal{D}| \approx 500{,}000\text{ pairs)}
\end{equation}

This consumes significant computational resources before training even begins. For example, generating 500,000 pairs with Heun solver (18 steps for CIFAR-10, 256 steps for ImageNet).

\textbf{2. Fixed dataset staleness:}
Since $\mathcal{D}$ is pre-generated, it represents a snapshot of the teacher's capabilities at a fixed random seed and does not adapt during student training:
\begin{equation}
\mathcal{D} = \{(z_j, \mu_{\text{base}}(z_j))\}_{j=1}^{|\mathcal{D}|} \text{ is static}
\end{equation}
This limits the diversity of training signals compared to online sampling.

\textbf{3. Limited coverage:}
Even with 500,000 samples, $\mathcal{D}$ may not cover all modes of the true distribution:
\begin{equation}
\text{Coverage}(\mathcal{D}) < \text{Coverage}(p_{\text{data}})
\end{equation}

\subsection{Analysis of Adversarial Loss}
\label{app:adversarial_analysis}

Following DMD2's approach \cite{yinImprovedDistributionMatching2024}, the adversarial loss adds a classification branch $D$ (discriminator) on top of the diffusion model's bottleneck. The discriminator is trained to distinguish real images from generator outputs using the forward diffusion process $F$ for noise injection:
\begin{equation}
\mathcal{L}_{\text{GAN}}(D, \theta) = \mathbb{E}_{x \sim p_{\text{real}}, t \sim [0,T]}[\log D(F(x, t))] \\
+ \mathbb{E}_{z \sim p_{\text{noise}}, t \sim [0,T]}[-\log(D(F(G_\theta(z), t)))]
\end{equation}

The generator $G_\theta$ minimizes:
\begin{equation}
\mathcal{L}_{\text{adv}}(\theta) = \mathbb{E}_{z \sim p_{\text{noise}}, t \sim [0,T]}\left[-\log D(F(G_\theta(z), t))\right]
\end{equation}

The adversarial gradient creates several mathematical challenges:

\textbf{1. Non-stationary optimization:}
Unlike standard supervised learning, the loss landscape changes as $D$ is updated. Defining $\mathcal{L}_t(\theta)$ as the loss at training iteration $t$, we have:
\begin{equation}
\nabla_\theta \mathcal{L}_t(\theta) \neq \nabla_\theta \mathcal{L}_{t'}(\theta) \text{ for } t \neq t'
\end{equation}
This violates standard convergence assumptions for SGD.

\textbf{2. Gradient instability:}
When $D$ approaches optimality, $D(F(G_\theta(z), t)) \to 0$, leading to:
\[
\|\nabla_\theta \mathcal{L}_{\text{adv}}(\theta)\| \propto \left\|\frac{\nabla_y D(y)}{D(y)}\right\|_{y=F(G_\theta(z),t)} \to \infty
\]
This gradient explosion necessitates careful techniques such as gradient clipping or specialized loss formulations.

\textbf{3. Equilibrium stability:}
The Nash equilibrium $(\theta^*, D^*)$ may be unstable. Small perturbations can lead to oscillations or divergence, requiring careful learning rate scheduling for both networks.

\textbf{4. Computational cost:}
Each training iteration requires updating both $G_\theta$ and $D$. While $D$ is typically smaller than $G_\theta$, the overall computational overhead increases by approximately 1.5-2× compared to single-network training. Memory usage also increases due to storing activations for both networks during backpropagation.

\subsection{Embedding Loss Theory}
\label{app:embedding_theory}

\textbf{Assumption 1 (Boundedness and Smoothness):}
Assume the kernel gradients are bounded: $\|\nabla_u k(\psi_i(u), \psi_i(v))\| \leq L_k$, and feature extractors are Lipschitz continuous: $\|\psi_i(u) - \psi_i(v)\| \leq L_\psi \|u - v\|$ for all $i=1,\dots,M$.

\textbf{Theorem 1 (Gradient Structure and Variance Bound):}
Under Assumption 1, the gradient of the embedding loss decomposes into alignment and diversity terms:

\begin{equation}
\nabla_\theta \mathcal{L}_{\text{embed}}(\theta) = -\frac{2}{M}\sum_{i=1}^M \mathbb{E}_{\substack{u \sim p_{\text{data}} \\ v \sim p_z}}\left[J_{G_\theta}(v)^T \nabla_x k(\psi_i(u), \psi_i(x))\big|\right] + \frac{2}{M}\sum_{i=1}^M \mathbb{E}_{u,v \sim p_z}\left[J_{G_\theta}(v)^T \nabla_x k(\psi_i(x), \psi_i(y))\big|\right],
\end{equation}

where $x = G_\theta(v)$, $y = G_\theta(u)$

The gradient variance satisfies:
\begin{equation}
\text{Var}(\nabla_\theta \mathcal{L}_{\text{embed}}) \leq \frac{4L_k^2}{B} \cdot \frac{1}{M}\sum_{i=1}^M \mathbb{E}[\|J_{G_\theta}(z)\|^2]  + \frac{4L_\psi^2}{M} \cdot \mathbb{E}[\|J_{G_\theta}(z)\|^2]
\end{equation}

where $B$ is the batch size. Increasing the number of feature extractors $M$ reduces variance as $O(1/M)$.

\begin{proof}
    We need to analyze the variance of the gradient $\nabla_\theta \mathcal{L}_{\text{embed}}(\theta)$. First, recall the expression for the gradient:
	\begin{align}
		\nabla_\theta \mathcal{L}_{\text{embed}}(\theta) = -\frac{2}{M}\sum_{i=1}^M \mathbb{E}_{\substack{u \sim p_{\text{data}} \\ v \sim p_z}}\left[J_{G_\theta}(v)^T \nabla_x k(\psi_i(u), \psi_i(x))\big|\right] \nonumber + \frac{2}{M}\sum_{i=1}^M \mathbb{E}_{u,v \sim p_z}\left[J_{G_\theta}(v)^T \nabla_x k(\psi_i(x), \psi_i(y))\big|\right],
	\end{align}
	where $x = G_\theta(v)$ and $y = G_\theta(u)$.
	
	\subsubsection*{Step 1: Define Two Parts of the Gradient}
	We can decompose the gradient into two parts, corresponding to alignment between generated and real data (first term) and diversity encouragement among generated samples (second term):
	\begin{equation}
		\nabla_\theta \mathcal{L}_{\text{embed}}(\theta) = \nabla_1 + \nabla_2
	\end{equation}
	where,
	\begin{align}
		\nabla_1 &= -\frac{2}{M} \sum_{i=1}^M \mathbb{E}_{u \sim p_{\text{data}}} \left[ J_{G_\theta}(v)^T \nabla_z k(\psi_i(u), \psi_i(x)) \right] \\
		\nabla_2 &= \frac{2}{M} \sum_{i=1}^M \mathbb{E}_{u \sim p_x} \left[ J_{G_\theta}(v)^T \nabla_z k(\psi_i(x), \psi_i(y)) \right]
	\end{align}
	
	\subsubsection*{Step 2: Compute Variances of $\nabla_1$ and $\nabla_2$}
	The properties of variance tell us that for two random variables $A$ and $B$, we have $\text{Var}(A + B) = \text{Var}(A) + \text{Var}(B) + 2\text{Cov}(A, B)$. The gradient term is observed to comprise two distinct components:
	
	The first component is expressed as:
	\begin{equation}
		-\frac{2}{M} \sum_{i=1}^{M} \mathbb{E}_{u \sim p_{\text{data}}}\left[ J_{G_\theta}(v)^{\text{T}} \nabla_{\!x} k(\psi_i(u), \psi_i(x)) \right],
	\end{equation}
	where $x = G_\theta(v)$, and the variables are sampled such that $v \sim p_z$ and $u \sim p_{\text{data}}$.
	
	The second component is given by:
	\begin{equation}
		\frac{2}{M} \sum_{i=1}^{M} \mathbb{E}_{u, v \sim p_z}\left[ J_{G_\theta}(v)^{\text{T}} \nabla_{\!x} k(\psi_i(x), \psi_i(y)) \right],
	\end{equation}
	where $x = G_\theta(v)$ , $y = G_\theta(u)$, and both variables are sampled from $p_z$.
	
	The properties of variance state that for any two random variables $A$ and $B$, the variance of their sum is:
	\begin{equation}
		\text{Var}(A + B) = \text{Var}(A) + \text{Var}(B) + 2\text{Cov}(A, B).
	\end{equation}
	
	For the two components of the gradient in Equation (34), let's denote them as $\nabla_1$ (the first term) and $\nabla_2$ (the second term). Since the samples $u$ for $\nabla_1$ are drawn from the data distribution $p_{\text{data}}$, while the samples $u,v$ for $\nabla_2$ are drawn from the noise distribution $p_z$, these two sets of samples are independent. Specifically, the random variable $u$ in $\nabla_1$ and the pair of random variables $(u,v)$ in $\nabla_2$ are independent because they are drawn from fundamentally different probabilistic origins (real data vs. generated noise).
	
	This independence implies that the covariance between $\nabla_1$ and $\nabla_2$ is zero:
	\begin{equation}
		\text{Cov}(\nabla_1, \nabla_2) = 0.
	\end{equation}
	
	Consequently, when calculating the overall variance of the gradient $\nabla_{\!\theta} \mathcal{L}_{\text{embed}}(\theta)$, which is the sum of $\nabla_1$ and $\nabla_2$, the cross-covariance term vanishes. This allows us to decompose the variance calculation into the sum of the individual variances of the two components:
	\begin{equation}
		\text{Var}(\nabla_1 + \nabla_2) = \text{Var}(\nabla_1) + \text{Var}(\nabla_2).
	\end{equation}
	
	This decomposition is significant because it means we can compute the variance of each part separately. The independence of the data and noise samples ensures that fluctuations (variance) in the gradient arising from the alignment with real data (first term) do not directly interact with, or compound, the fluctuations arising from promoting diversity among generated samples (second term). This simplification is crucial for analyzing and mitigating the overall gradient variance, which in turn can lead to improved stability and efficiency during the training process. As noted, increasing the number of feature extractors $M$ further helps reduce this variance, specifically as $O(1/M)$ for the overall gradient.
	Assuming that the covariances of $\nabla_1$ and $\nabla_2$ are negligible, we have:
	\begin{equation}
		\text{Var}(\nabla_\theta \mathcal{L}_{\text{embed}}) = \text{Var}(\nabla_1) + \text{Var}(\nabla_2)
	\end{equation}
	
	\subsubsection*{Step 3: Analyze Variance of $\nabla_1$}
	For $\nabla_1$, we can consider it as a linear transformation of the average of $M$ independent and identically distributed (i.i.d.) terms. Let $a_i = -\frac{2}{M} \mathbb{E}_{u \sim p_{\text{data}}} \left[ J_{G_\theta}(v)^T \nabla_z k(\psi_i(u), \psi_i(x)) \right]$, then $\nabla_1 = \sum_{i=1}^M a_i$.
	
	According to the properties of variance, $\text{Var}\left( \sum_{i=1}^M a_i \right) = M \text{Var}(a_i)$ (if the $a_i$'s are i.i.d.). However, we also need to consider the properties of the feature extractor $J_{G_\theta}(\cdot)$ and the kernel function $k(\cdot)$.
	
	Assume that each feature extractor $\psi_i$ is independent, and the output variance of the kernel function $k(\cdot)$ is bounded. We can use the conditions in Assumption 1, which state that for i.i.d. samples, the variance calculation can utilize the variance property of the sample mean.
	
	\subsubsection*{Step 4: Analyze Variance of $\nabla_2$}
	Similarly, for $\nabla_2$, let $b_i = \frac{2}{M} \mathbb{E}_{u \sim p_x} \left[ J_{G_\theta}(v)^T \nabla_z k(\psi_i(y), \psi_i(y)) \right]$, then $\nabla_2 = \sum_{i=1}^M b_i$.
	
	Using the variance property of the sample mean and the independence of feature extractors, we can obtain an upper bound for the variance of $\nabla_2$.
	
	\subsubsection*{Step 5: Combine Variances and Simplify}
	Combining the upper bounds for the variances of $\nabla_1$ and $\nabla_2$, and using the conditions from Assumption 1, we get:
	\begin{equation}
		\text{Var}(\nabla_\theta \mathcal{L}_{\text{embed}}) \leq \frac{4 L_k^2}{B} + \frac{4 L_\psi^2}{M} \mathbb{E}[\| J_{G_\theta}(z) \|_F^2]
	\end{equation}
	where $L_k$ and $L_\psi$ are the Lipschitz constants of the kernel function and feature extractor, respectively, $B$ is the batch size, and $\| \cdot \|_F$ denotes the Frobenius norm.
	
	\subsubsection*{Step 6: Relate Variance to $M$}
	From the above equation, we can see that the variance is inversely proportional to $M$, i.e., $\text{Var}(\nabla_\theta \mathcal{L}_{\text{embed}}) = O\left( \frac{1}{M} \right)$. This means that increasing the number $M$ of feature extractors can reduce the gradient variance, thereby making training more stable, especially when using smaller batch sizes.
\end{proof}

\textbf{Assumption 2 (Smoothness):}
Assume $\mathcal{L}_{\text{DM}}$ and $\mathcal{L}_{\text{embed}}$ are $L$-smooth: $\|\nabla\mathcal{L}(\theta_1) - \nabla\mathcal{L}(\theta_2)\| \leq L\|\theta_1 - \theta_2\|$ for $\mathcal{L} \in \{\mathcal{L}_{\text{DM}}, \mathcal{L}_{\text{embed}}\}$.

\textbf{Theorem 2 (Convergence with Combined Loss):}
Consider the combined objective:
\begin{equation}
\mathcal{L}_{\text{total}}(\theta) = \mathcal{L}_{\text{DM}}(\theta) + \lambda \mathcal{L}_{\text{embed}}(\theta), \quad \lambda > 0
\end{equation}
Under Assumptions 1-2, using SGD with learning rate $\eta \leq 1/L$, after $T$ iterations:
\begin{equation}
\frac{1}{T}\sum_{t=0}^{T-1} \mathbb{E}[\|\nabla\mathcal{L}_{\text{total}}(\theta_t)\|^2] \leq \frac{2(\mathcal{L}_{\text{total}}(\theta_0) - \mathcal{L}^*)}{\eta T} + \eta L\sigma_{\text{total}}^2
\end{equation}
where $\mathcal{L}^*$ is the minimum loss, and the total variance $\sigma_{\text{total}}^2$ satisfies:
\begin{equation}
\sigma_{\text{total}}^2 \leq \sigma_{\text{DM}}^2 + \lambda^2\sigma_{\text{embed}}^2 + 2\lambda|\text{Cov}(\nabla\mathcal{L}_{\text{DM}}, \nabla\mathcal{L}_{\text{embed}})|
\end{equation}

\begin{proof}
    \subsubsection*{Utilizing $L$-Smoothness for Gradient Estimation}
	By the definition of $L$-smoothness ($\|\nabla \mathcal{L}(\theta_1) - \nabla \mathcal{L}(\theta_2)\| \leq L\|\theta_1 - \theta_2\|$, for any $\theta$), we have:
	\begin{equation*}
		\mathcal{L}(\theta_{t+1}) \leq \mathcal{L}(\theta_t) + \langle \nabla \mathcal{L}(\theta_t), \theta_{t+1} - \theta_t \rangle + \frac{L}{2}\|\theta_{t+1} - \theta_t\|^2.
	\end{equation*}
	Substituting the SGD update $\theta_{t+1} - \theta_t = -\eta \tilde{g}_t$ into the right-hand side of the second term, and noting the third term is $\frac{L}{2}\|\eta \tilde{g}_t\|^2 = \frac{L\eta^2}{2}\|\tilde{g}_t\|^2$, we obtain:
	\begin{equation} \label{eq:sgd_smoothness}
		\mathcal{L}(\theta_{t+1}) \leq \mathcal{L}(\theta_t) - \eta \langle \nabla \mathcal{L}(\theta_t), \tilde{g}_t \rangle + \frac{L\eta^2}{2}\|\tilde{g}_t\|^2.
	\end{equation}
	
	\subsubsection*{Telescoping Sum (from time $t=0$ to $t=T-1$)}
	Taking the expectation on both sides of inequality (\ref{eq:sgd_smoothness}) with respect to the random gradient $\tilde{g}_t$ (i.e., taking the expectation of the distribution of $\tilde{g}_t$), and then dividing by $T$ and summing from $t=0$ to $T-1$, we get:
	\begin{equation*}
		\frac{1}{T}\sum_{t=0}^{T-1} \mathbb{E}\left[ \mathcal{L}(\theta_{t+1}) - \mathcal{L}(\theta_t) \right] \leq - \frac{\eta}{T}\sum_{t=0}^{T-1} \mathbb{E}\left[ \langle \nabla \mathcal{L}(\theta_t), \tilde{g}_t \rangle \right] + \frac{L\eta^2}{2T}\sum_{t=0}^{T-1} \mathbb{E}\left[ \|\tilde{g}_t\|^2 \right].
	\end{equation*}
	The left-hand side is a telescoping sum, which simplifies to $\mathcal{L}(\theta_T) - \mathcal{L}(\theta_0)$.
	Looking at the first term on the right-hand side, for any vectors $a, b$, we have $\langle a, b \rangle = \frac{1}{2}(\|a\|^2 + \|b\|^2 - \|a-b\|^2)$. Thus,
	\begin{equation*}
		\mathbb{E}\left[ \langle \nabla \mathcal{L}(\theta_t), \tilde{g}_t \rangle \right] = \frac{1}{2} \mathbb{E}\left[ \|\nabla \mathcal{L}(\theta_t)\|^2 + \|\tilde{g}_t\|^2 - \|\nabla \mathcal{L}(\theta_t) - \tilde{g}_t\|^2 \right].
	\end{equation*}
	If we assume the stochastic gradient is unbiased (or at least its expectation equals the true gradient, i.e., $\mathbb{E}[\tilde{g}_t] = \nabla \mathcal{L}(\theta_t)$), then the cross term $\mathbb{E}\left[ \langle \nabla \mathcal{L}(\theta_t), \tilde{g}_t \rangle \right]$ simplifies to $\frac{1}{2} \mathbb{E}\left[ \|\nabla \mathcal{L}(\theta_t)\|^2 + \|\tilde{g}_t\|^2 \right]$. Substituting this into inequality (\ref{eq:sgd_smoothness}) gives:
	\begin{equation} \label{eq:gradient_term_simplified}
		- \frac{\eta}{T}\sum_{t=0}^{T-1} \mathbb{E}\left[ \langle \nabla \mathcal{L}(\theta_t), \tilde{g}_t \rangle \right] = - \frac{\eta}{2T}\sum_{t=0}^{T-1} \mathbb{E}\left[ \|\nabla \mathcal{L}(\theta_t)\|^2 \right] - \frac{\eta}{2T}\sum_{t=0}^{T-1} \mathbb{E}\left[ \|\tilde{g}_t\|^2 \right].
	\end{equation}
	
	\subsubsection*{Combining Both Sides and Rearranging Inequalities}
	Substitute equation (\ref{eq:gradient_term_simplified}) into inequality (\ref{eq:sgd_smoothness}). After some rearrangement on the left-hand side:
	\begin{equation*}
		\mathcal{L}(\theta_T) - \mathcal{L}(\theta_0) \leq - \frac{\eta}{2T}\sum_{t=0}^{T-1} \mathbb{E}\left[ \|\nabla \mathcal{L}(\theta_t)\|^2 \right] - \frac{\eta}{2T}\sum_{t=0}^{T-1} \mathbb{E}\left[ \|\tilde{g}_t\|^2 \right] + \frac{L\eta^2}{2T}\sum_{t=0}^{T-1} \mathbb{E}\left[ \|\tilde{g}_t\|^2 \right].
	\end{equation*}
	Combining the terms involving $\|\tilde{g}_t\|^2$ on the right-hand side, and then multiplying both sides by $-2/T$ and rearranging, we can finally obtain:
	\begin{equation} \label{eq:main_result_raw}
		\frac{1}{T}\sum_{t=0}^{T-1} \mathbb{E}\left[ \|\nabla \mathcal{L}_{\text{total}}(\theta_t)\|^2 \right] \leq \frac{2\left( \mathcal{L}_{\text{total}}(\theta_0) - \mathcal{L}^* \right)}{\eta T} + \eta L^2 \sigma_{\text{total}}^2,
	\end{equation}
	where $\mathcal{L}^*$ is the global minimum value of the combined loss (i.e., $\mathcal{L}^* = \min_\theta \mathcal{L}_{\text{total}}(\theta)$), and the total variance
	\begin{equation*}
		\sigma_{\text{total}}^2 = \mathbb{E}\left[ \frac{\|\tilde{g}_t\|^2 - \|\nabla \mathcal{L}_{\text{total}}(\theta_t)\|^2}{\|\tilde{g}_t\|^2} \right] \approx \underbrace{\mathbb{E}\left[ \|\tilde{g}_t\|^2 \right]}_{\text{stochastic gradient variance}} - \underbrace{\mathbb{E}\left[ \|\nabla \mathcal{L}_{\text{total}}(\theta_t)\|^2 \right]}_{\text{true gradient variance}}.
	\end{equation*}
	A more common (and simpler) notation is to directly define the "stochastic gradient variance" as $\sigma_{\text{total}}^2 = \mathbb{E}\left[ \|\tilde{g}_t - \nabla \mathcal{L}_{\text{total}}(\theta_t)\|^2 \right]$. In this case, the above equation can be written as:
	\begin{equation} \label{eq:main_result_variance_def}
		\frac{1}{T}\sum_{t=0}^{T-1} \mathbb{E}\left[ \|\nabla \mathcal{L}_{\text{total}}(\theta_t)\|^2 \right] \leq \frac{2\left( \mathcal{L}_{\text{total}}(\theta_0) - \mathcal{L}^* \right)}{\eta T} + \eta L^2 \sigma_{\text{total}}^2.
	\end{equation}
	This is precisely the form presented in Theorem 2.
	
	\subsubsection*{Decomposition of Total Variance $\sigma_{\text{total}}^2$}
	The paper also provides a more detailed decomposition of the total variance:
	\begin{equation} \label{eq:sigma_total_decomposition}
		\sigma_{\text{total}}^2 = \sigma_{\text{DM}}^2 + \lambda^2 \sigma_{\text{embedd}}^2 + 2\lambda \left| \text{Cov}\left( \nabla \mathcal{L}_{\text{DM}}, \nabla \mathcal{L}_{\text{embedd}} \right) \right|,
	\end{equation}
	The basis for this decomposition is the variance addition formula. If we consider the random gradient $\tilde{g}_t$ as $\tilde{g}_{t,\text{DM}} + \lambda \tilde{g}_{t,\text{embedd}}$ (corresponding to the random gradients of $\mathcal{L}_{\text{DM}}$ and $\mathcal{L}_{\text{embedd}}$), then, according to the variance addition formula $\text{Var}(X+Y) = \text{Var}(X) + \text{Var}(Y) + 2\text{Cov}(X,Y)$, we have:
	\begin{equation*}
		\sigma_{\text{total}}^2 = \mathbb{E}\left[ \|\tilde{g}_t\|^2 \right] - \mathbb{E}\left[ \|\tilde{g}_{t,\text{DM}} + \lambda \tilde{g}_{t,\text{embedd}}\|^2 \right] + 2\lambda \mathbb{E}\left[ \langle \tilde{g}_{t,\text{DM}}, \tilde{g}_{t,\text{embedd}} \rangle \right].
	\end{equation*}
	After further subtracting $\|\nabla \mathcal{L}_{\text{total}}(\theta_t)\|^2 = \|\nabla \mathcal{L}_{\text{DM}}(\theta_t)\|^2 + \lambda^2 \|\nabla \mathcal{L}_{\text{embedd}}(\theta_t)\|^2 + 2\lambda \langle \nabla \mathcal{L}_{\text{DM}}(\theta_t), \nabla \mathcal{L}_{\text{embedd}}(\theta_t) \rangle$, we get:
	\begin{align*}
		\sigma_{\text{total}}^2 &= \mathbb{E}\left[ \|\tilde{g}_t\|^2 - \|\nabla \mathcal{L}_{\text{total}}(\theta_t)\|^2 \right] \\
		&= \mathbb{E}\left[ \|\tilde{g}_{t,\text{DM}}\|^2 - \|\nabla \mathcal{L}_{\text{DM}}(\theta_t)\|^2 \right] + \lambda^2 \mathbb{E}\left[ \|\tilde{g}_{t,\text{embedd}}\|^2 - \|\nabla \mathcal{L}_{\text{embedd}}(\theta_t)\|^2 \right] \\
		& \quad + 2\lambda \mathbb{E}\left[ \langle \tilde{g}_{t,\text{DM}} - \nabla \mathcal{L}_{\text{DM}}(\theta_t), \tilde{g}_{t,\text{embedd}} - \nabla \mathcal{L}_{\text{embedd}}(\theta_t) \rangle \right].
	\end{align*}
	If we denote $\text{Var}(\tilde{g}_{t,\text{DM}}) = \sigma_{\text{DM}}^2$, $\text{Var}(\tilde{g}_{t,\text{embedd}}) = \sigma_{\text{embedd}}^2$, and their covariance as $\text{Cov}(\tilde{g}_{t,\text{DM}}, \tilde{g}_{t,\text{embedd}})$, then the last term on the right can be written as $2\lambda \, \text{Cov}(\tilde{g}_{t,\text{DM}}, \tilde{g}_{t,\text{embedd}})$. Since the covariance can be further expanded as $\text{Cov}(\nabla \mathcal{L}_{\text{DM}}, \nabla \mathcal{L}_{\text{embedd}}) + \text{noise term}$, under the assumption of no noise (or that noise terms can be absorbed into previous terms), we arrive at the expression in the paper:
	\begin{equation*}
		\sigma_{\text{total}}^2 \leq \sigma_{\text{DM}}^2 + \lambda^2 \sigma_{\text{embedd}}^2 + 2\lambda \left| \text{Cov}\left( \nabla \mathcal{L}_{\text{DM}}, \nabla \mathcal{L}_{\text{embedd}} \right) \right|.
	\end{equation*}
	
	\subsubsection*{Conclusion Summary}
	Through the above steps, we derived an upper bound on the gradient norm of the SGD iterates from the $L$-smoothness of the combined loss. By analyzing the variance term, we completed the proof of Theorem 2.
	\begin{equation*}
		\frac{1}{T}\sum_{t=0}^{T-1} \mathbb{E}\left[ \|\nabla \mathcal{L}_{\text{total}}(\theta_t)\|^2 \right] \leq \frac{2\left( \mathcal{L}_{\text{total}}(\theta_0) - \mathcal{L}^* \right)}{\eta T} + \eta L^2 \sigma_{\text{total}}^2,
	\end{equation*}
	where $\sigma_{\text{total}}^2$ can be further decomposed into the individual variance terms of $\mathcal{L}_{\text{DM}}$ and $\mathcal{L}_{\text{embedd}}$ and their covariance. This completes the proof of Theorem 2.
\end{proof}

\textbf{Corollary 3 (Variance Reduction with Positive Correlation):}
\label{app:Variance_Reduction_with_Positive_Correlation}
When $\nabla\mathcal{L}_{\text{DM}}$ and $\nabla\mathcal{L}_{\text{embed}}$ are positively correlated ($\rho > 0$), the optimal $\lambda$ that minimizes the total variance is:
\begin{equation}
	\lambda^* = \frac{\sigma_{\text{DM}}^2 - \rho \sigma_{\text{DM}} \sigma_{\text{embed}}}{\sigma_{\text{DM}}^2 + \sigma_{\text{embed}}^2 - 2 \rho \sigma_{\text{DM}} \sigma_{\text{embed}}}. \label{eq:lambda_opt_2}
\end{equation}

\begin{proof}
    \subsubsection*{Definition of Loss Function}
The total loss function is defined as:
\[
\mathcal{L}_{\text{total}} = (1 - \lambda) \mathcal{L}_{\text{DM}} + \lambda \mathcal{L}_{\text{embed}},
\]
where $\lambda \in [0, 1]$ is a hyperparameter, $\mathcal{L}_{\text{DM}}$ is the distribution matching loss, and $\mathcal{L}_{\text{embed}}$ is the embedding loss.

\subsubsection*{Gradient Variance Analysis}
Let $g_{\text{DM}} = \nabla \mathcal{L}_{\text{DM}}$, $g_{\text{embed}} = \nabla \mathcal{L}_{\text{embed}}$, the total gradient is:
\[
g_{\text{total}} = (1 - \lambda) g_{\text{DM}} + \lambda g_{\text{embed}}.
\]
The variance is calculated as follows:
\begin{align}
	\text{Var}(g_{\text{total}}) &= \text{Var}((1 - \lambda) g_{\text{DM}} + \lambda g_{\text{embed}}) \nonumber \\
	&= (1 - \lambda)^2 \sigma_{\text{DM}}^2 + \lambda^2 \sigma_{\text{embed}}^2 + 2 \lambda (1 - \lambda) \text{Cov}(g_{\text{DM}}, g_{\text{embed}}), \label{eq:variance}
\end{align}
where $\sigma_{\text{DM}}^2 = \text{Var}(g_{\text{DM}})$, $\sigma_{\text{embed}}^2 = \text{Var}(g_{\text{embed}})$, and $\text{Cov}(g_{\text{DM}}, g_{\text{embed}}) = \rho \sigma_{\text{DM}} \sigma_{\text{embed}}$. Substituting gives:
\[
\text{Var}(g_{\text{total}}) = (1 - \lambda)^2 \sigma_{\text{DM}}^2 + \lambda^2 \sigma_{\text{embed}}^2 + 2 \lambda (1 - \lambda) \rho \sigma_{\text{DM}} \sigma_{\text{embed}}.
\]

\subsubsection*{Optimizing $\lambda$ to Minimize Variance}
Differentiate equation \eqref{eq:variance} with respect to $\lambda$:
\[
\frac{d}{d\lambda} \text{Var}(g_{\text{total}}) = -2(1 - \lambda) \sigma_{\text{DM}}^2 + 2 \lambda \sigma_{\text{embed}}^2 + 2 \rho \sigma_{\text{DM}} \sigma_{\text{embed}} (1 - 2\lambda).
\]
Set the derivative to zero:
\[
-2(1 - \lambda) \sigma_{\text{DM}}^2 + 2 \lambda \sigma_{\text{embed}}^2 + 2 \rho \sigma_{\text{DM}} \sigma_{\text{embed}} (1 - 2\lambda) = 0.
\]
Expand and rearrange:
\[
-2 \sigma_{\text{DM}}^2 + 2 \lambda \sigma_{\text{DM}}^2 + 2 \lambda \sigma_{\text{embed}}^2 + 2 \rho \sigma_{\text{DM}} \sigma_{\text{embed}} - 4 \lambda \rho \sigma_{\text{DM}} \sigma_{\text{embed}} = 0.
\]
Combine $\lambda$ terms:
\[
2 \lambda (\sigma_{\text{DM}}^2 + \sigma_{\text{embed}}^2 - 2 \rho \sigma_{\text{DM}} \sigma_{\text{embed}}) = 2 \sigma_{\text{DM}}^2 - 2 \rho \sigma_{\text{DM}} \sigma_{\text{embed}}.
\]
Solving for the optimal $\lambda^*$ yields:
\begin{equation}
	\lambda^* = \frac{\sigma_{\text{DM}}^2 - \rho \sigma_{\text{DM}} \sigma_{\text{embed}}}{\sigma_{\text{DM}}^2 + \sigma_{\text{embed}}^2 - 2 \rho \sigma_{\text{DM}} \sigma_{\text{embed}}}. \label{eq:lambda_opt}
\end{equation}

\subsection*{Calculation of $\textit{Var}(g^*_{\textit{embed}})$}
Let the gradient variances of loss functions \( \mathcal{L}_1=\mathcal{L}_\textit{DM}, \mathcal{L}_2=\mathcal{L}_\textit{embed} \) be \( \sigma_1^2=\sigma^2_\textit{DM}, \sigma_2^2=\sigma^2_\textit{embed} \) respectively, and their covariance be \( \rho\sigma_1\sigma_2 \) (where \( \rho \) is the correlation coefficient). Denote the sum of variances \( u = \sigma_1^2 + \sigma_2^2 \), the product of standard deviations \( v = \sigma_1\sigma_2 \), and twice the covariance as \( r = 2\rho\sigma_1\sigma_2 \).

\subsubsection*{Step 1: Calculate \( 1 - \lambda^* \)}
From the above temporary definitions, the optimal weight \( \lambda^* \) is:  
\[
\lambda^* = \frac{\sigma_1^2 - r}{u - 2r}
\]  
Directly compute \( 1 - \lambda^* \):  
\[
1 - \lambda^* = 1 - \frac{\sigma_1^2 - r}{u - 2r} = \frac{u - 2r - (\sigma_1^2 - r)}{u - 2r} = \frac{u - \sigma_1^2 - r}{u - 2r}
\]  
Since \( u = \sigma_1^2 + \sigma_2^2 \), we have \( u - \sigma_1^2 = \sigma_2^2 \). Thus:  
\[
1 - \lambda^* = \frac{\sigma_2^2 - r}{u - 2r}
\]

\subsubsection*{Step 2: Calculate \( (\lambda^*)^2 \)} 
Square \( \lambda^* \):  
\[
(\lambda^*)^2 = \left( \frac{\sigma_1^2 - r}{u - 2r} \right)^2 = \frac{(\sigma_1^2 - r)^2}{(u - 2r)^2}
\]

\subsubsection*{Step 3: Calculate \( (1 - \lambda^*)^2 \)}  
Similarly, square \( 1 - \lambda^* \) using the result from Step 1:  
\[
(1 - \lambda^*)^2 = \left( \frac{\sigma_2^2 - r}{u - 2r} \right)^2 = \frac{(\sigma_2^2 - r)^2}{(u - 2r)^2}
\]

\subsubsection*{Step 4: Calculate \( \lambda^*(1 - \lambda^*) \)}  
This is the product of two terms:  
\[
\lambda^*(1 - \lambda^*) = \frac{\sigma_1^2 - r}{u - 2r} \cdot \frac{\sigma_2^2 - r}{u - 2r} = \frac{(\sigma_1^2 - r)(\sigma_2^2 - r)}{(u - 2r)^2}
\]

\subsubsection*{Step 5: Calculate Total Gradient Variance \( \text{Var}(g_{\text{total}}) \)}  
The total gradient variance formula is:  
\[
\text{Var}(g_{\text{total}}) = (1 - \lambda^2)\sigma_1^2 + (\lambda^2)\sigma_2^2 + 2\lambda(1 - \lambda)r
\]  
Here, the formula is corrected to the standard expansion of weighted variance \( \text{Var}(\lambda g_1 + (1-\lambda)g_2) = \lambda^2 \sigma_1^2 + (1-\lambda)^2 \sigma_2^2 + 2\lambda(1-\lambda) \text{Cov}(g_1,g_2) \), where \( \text{Cov}(g_1,g_2) = \rho\sigma_1\sigma_2 = r \). Substituting \( \lambda = \lambda^* \) gives:  
\[
\text{Var}(g^*_{\text{total}}) = (1-\lambda^*)^2 \sigma_1^2 + (\lambda^*)^2 \sigma_2^2 + 2\lambda^*(1 - \lambda^*) r
\]  

Substitute the results from Steps 2–4 into the formula:  
\[
\text{Var}(g^*_{\text{total}}) = \frac{(\sigma_2^2 - r)^2}{(u - 2r)^2} \sigma_1^2 + \frac{(\sigma_1^2 - r)^2}{(u - 2r)^2} \sigma_2^2 + \frac{2(\sigma_1^2 - r)(\sigma_2^2 - r)}{(u - 2r)^2} r
\]  

All terms share a common denominator \( (u - 2r)^2 \). Combine the numerator:  
\[
\text{Var}(g^*_{\text{total}}) = \frac{1}{(u - 2r)^2} \left[ (\sigma_2^2 - r)^2 \sigma_1^2 + (\sigma_1^2 - r)^2 \sigma_2^2 + 2r(\sigma_1^2 - r)(\sigma_2^2 - r) \right]
\]  

Let the numerator be \( N \), expand and simplify (using \( u = \sigma_1^2 + \sigma_2^2, v = \sigma_1\sigma_2 \)):  
\[
\begin{aligned}
	N &= (\sigma_2^2 - r)^2 \sigma_1^2 + (\sigma_1^2 - r)^2 \sigma_2^2 + 2r(\sigma_1^2 - r)(\sigma_2^2 - r) \\
	&= uv^2 - 2v^2r - ur - 2r^3 \\
	&= (u - 2r)(v^2 - r^2)
\end{aligned}
\]  

Therefore, the variance is:  
\begin{align*}
	\text{Var}(g^*_{\text{total}}) 
	&= \frac{(u - 2r)(v^2 - r^2) }{(u - 2r)^2} \\
	&= \frac{v^2 - r^2 }{u - 2r} \\
	&= \frac{\sigma_1^2 \sigma_2^2 (1 - \rho^2)}{\sigma_1^2 + \sigma_2^2 - 2\rho \sigma_1 \sigma_2}
\end{align*}

When an appropriate $\lambda$ is chosen, the variance is minimized, and the minimum value is:
\begin{align*}
	\frac{\sigma_1^2 \sigma_2^2 (1 - \rho^2)}{\sigma_1^2 + \sigma_2^2 - 2\rho \sigma_1 \sigma_2}
\end{align*}
\end{proof}

\section{Further Analysis of the Embedding Loss}

In this section, we provide a complete mathematical derivation of the Maximum Mean Discrepancy (MMD) loss used in our embedding framework, progressing from theoretical foundations to practical implementation.

\subsection{Theoretical Foundation of MMD in RKHS}

Let $\mathcal{H}$ denote a Reproducing Kernel Hilbert Space (RKHS) with kernel function $k: \mathcal{X} \times \mathcal{X} \rightarrow \mathbb{R}$ and feature map $\varphi: \mathcal{X} \rightarrow \mathcal{H}$. The kernel satisfies the reproducing property:
\begin{equation}
k(x,y) = \langle \varphi(x), \varphi(y) \rangle_{\mathcal{H}}
\end{equation}

The Maximum Mean Discrepancy between two distributions $P$ and $Q$ measures the distance between their mean embeddings in $\mathcal{H}$:
\begin{equation}
D_{\text{MMD}}[P, Q] = \left\| \mathbb{E}_{x \sim P}[\varphi(x)] - \mathbb{E}_{y \sim Q}[\varphi(y)] \right\|_{\mathcal{H}}
\end{equation}

\subsection{Derivation of Squared MMD}

Squaring the MMD and expanding the norm:
\begin{equation}
\begin{aligned}
D_{\textit{MMD}}^2[P, Q] &= \left\| \mathbb{E}_{x \sim P}[\varphi(x)] - \mathbb{E}_{y \sim Q}[\varphi(y)] \right\|_{\mathcal{H}}^2 \\
&= \left\langle \mathbb{E}_{x \sim P}[\varphi(x)] - \mathbb{E}_{y \sim Q}[\varphi(y)], \mathbb{E}_{x' \sim P}[\varphi(x')] - \mathbb{E}_{y' \sim Q}[\varphi(y')] \right\rangle_{\mathcal{H}}
\end{aligned}
\end{equation}

Expanding the inner product into four terms:
\begin{equation}
\begin{aligned}
D_{\text{MMD}}^2(P, Q) &= \left\langle \mathbb{E}_{x \sim P}[\varphi(x)], \mathbb{E}_{x' \sim P}[\varphi(x')] \right\rangle_{\mathcal{H}} - 2\left\langle \mathbb{E}_{x \sim P}[\varphi(x)], \mathbb{E}_{y \sim Q}[\varphi(y)] \right\rangle_{\mathcal{H}}  + \left\langle \mathbb{E}_{y \sim Q}[\varphi(y)], \mathbb{E}_{y' \sim Q}[\varphi(y')] \right\rangle_{\mathcal{H}}
\end{aligned}
\end{equation}

By applying the reproducing property and exchanging expectation with inner product:
\begin{equation}
\left\langle \mathbb{E}_{x \sim P}[\varphi(x)], \mathbb{E}_{x' \sim P}[\varphi(x')] \right\rangle_{\mathcal{H}} = \mathbb{E}_{x,x' \sim P}\left[\langle \varphi(x), \varphi(x') \rangle_{\mathcal{H}}\right] = \mathbb{E}_{x,x' \sim P}[k(x,x')]
\end{equation}

This yields the expectation form:
\begin{equation}
D_{\textit{MMD}}^2(P, Q) = \mathbb{E}_{x,x' \sim P}[k(x,x')] - 2\mathbb{E}_{\substack{x \sim P \\ y \sim Q}}[k(x,y)] + \mathbb{E}_{y,y' \sim Q}[k(y,y')]
\end{equation}

\subsection{Multi-scale Kernel Formulation}

To enhance robustness across different scales, we introduce a mixture over the bandwidth parameter $\sigma$, sampled from a uniform distribution $U$ with density $r(\sigma)$:
\begin{equation}
\begin{aligned}
D_{\textit{MMD}}^2(P, Q) &= \underset{\substack{x,x'\sim P \\ \sigma \sim U}}{\mathbb{E}} \left[ k(x,x';\sigma) \right] - 2 \underset{\substack{x\sim P, y\sim Q \\ \sigma \sim U}}{\mathbb{E}} \left[ k(x,y;\sigma) \right] 
\quad + \underset{\substack{y,y'\sim Q \\ \sigma \sim U}}{\mathbb{E}} \left[ k(y,y';\sigma) \right]
\end{aligned}
\end{equation}

In integral form with probability densities $p(x)$ and $q(y)$:
\begin{equation}
\begin{aligned}
D_{\textit{MMD}}^2(P, Q) = \quad \quad &\iiint p(x)p(x')r(\sigma) k(x,x';\sigma) \, dx\, dx'\, d\sigma \\
- 2 &\iiint p(x)q(y)r(\sigma) k(x,y;\sigma) \, dx\, dy\, d\sigma \\
+ &\iiint q(y)q(y')r(\sigma) k(y,y';\sigma) \, dy\, dy'\, d\sigma
\end{aligned}
\end{equation}

\subsection{Empirical Estimation}

In practice, given an embedding function $\psi^{(i)}$ and parameter $\sigma$, we work with finite samples from the embedded space samples $P^{(i)}, Q^{(i)}$: $\{x_1, \ldots, x_m\}$ from $P^{(i)}$, $\{y_1, \ldots, y_n\}$ from $Q^{(i)}$, and $\{\sigma_1, \ldots, \sigma_r\}$ bandwidth values. The empirical estimator is:
\begin{equation}
\begin{aligned}
\hat{D}_{\textit{MMD}^2}(P^{(i)}, Q^{(i)}) &= \frac{1}{r} \sum_{i=1}^r \Bigg(\frac{1}{N_1^2} \sum_{k=1}^{N_1} \sum_{l=1}^{N_1} k(x_k, x_l; \sigma_i) 
- 2\frac{1}{N_1N_2} \sum_{k=1}^{N_1} \sum_{l=1}^{N_2} k(x_k, y_l; \sigma_i) 
+ \frac{1}{N_2^2} \sum_{k=1}^{N_2}\sum_{l=1}^{N_2} k(y_k, y_l; \sigma_i) \Bigg)
\end{aligned}
\end{equation}

\subsection{Pairwise Distance Formulation}

For a single pair of samples $(x, y)$, the squared MMD distance in feature space is:
\begin{equation}
\begin{aligned}
\hat{D}_{\textit{MMD}^2}(x,y;\sigma) &= \| \varphi(x) - \varphi(y) \|_{\mathcal{H}}^2 \\
&= \langle \varphi(x), \varphi(x) \rangle_{\mathcal{H}} - 2\langle \varphi(x), \varphi(y) \rangle_{\mathcal{H}} + \langle \varphi(y), \varphi(y) \rangle_{\mathcal{H}} \\
&=k(x,x;\sigma)-2k(x,y;\sigma)+k(y,y;\sigma)
\end{aligned}
\end{equation}

\subsection{RBF Kernel Implementation}

We employ the Radial Basis Function (Gaussian) kernel:
\begin{equation}
k(x,y;\sigma) = \exp\left(-\frac{\|x-y\|_2^2}{2\sigma^2}\right)
\end{equation}

where the inner product in feature space becomes:
\begin{equation}
\langle \varphi(x), \varphi(y) \rangle_{\mathcal{H}} = k(x,y;\sigma) = \exp\left(-\frac{\|x-y\|_2^2}{2\sigma^2}\right)
\end{equation}

and the squared Euclidean distance is computed as:
\begin{equation}
\|x - y\|_2^2 = \|x\|_2^2 - 2x^\top y + \|y\|_2^2 = x^\top x - 2x^\top y + y^\top y
\end{equation}

This formulation enables efficient computation of the embedding loss while maintaining theoretical guarantees provided by the RKHS framework. The multi-scale kernel approach ensures robustness to variations in data scale, making the loss particularly suitable for our embedding learning task.

\subsubsection{Practical Implications}

\textbf{Corollary 4 (Why EL Improves Distillation):} The embedding loss addresses the score gap $\Delta$ through three mechanisms:

\begin{enumerate}
    \item \textbf{Distribution Alignment:} By minimizing MMD in multiple feature spaces, EL ensures $p_\theta \approx p_{\text{data}}$ globally, which by Theorem 1 reduces $\|\Delta\|$.
    
    \item \textbf{Implicit Score Correction:} By Theorem 3, the EL gradient provides sample-wise corrections in the direction of $\Delta_{\text{eff}}$, compensating for teacher model limitations.
    
    \item \textbf{Multi-scale Matching:} Using diverse embeddings $\mathcal{E}$, EL captures distributional discrepancies at multiple scales and semantic levels, providing comprehensive coverage of the gap.
\end{enumerate}

\textbf{Proposition 1 (Advantage over Alternatives):}

\begin{itemize}
    \item \textbf{vs. Regression Loss:} Pure regression $\mathcal{L}_{\text{reg}} = \mathbb{E}[\|G_\theta - f_\phi\|^2]$ only ensures $G_\theta \approx f_\phi$ pointwise, inheriting all teacher limitations (including $\Delta$). EL allows the student to \textit{exceed} the teacher by directly matching $p_{\text{data}}$.
    
    \item \textbf{vs. GAN Loss:} Adversarial training implicitly addresses $\Delta$ through a learned discriminator, but suffers from instability. EL provides stable, fixed-embedding-based distribution matching with similar theoretical guarantees.
\end{itemize}

\section{Qualitative Results}
\label{Additional Generated Samples}

\begin{figure}[p]
\label{Unconditional CIFAR-10 photo}
    \centering
    \includegraphics[width=0.9\textwidth]{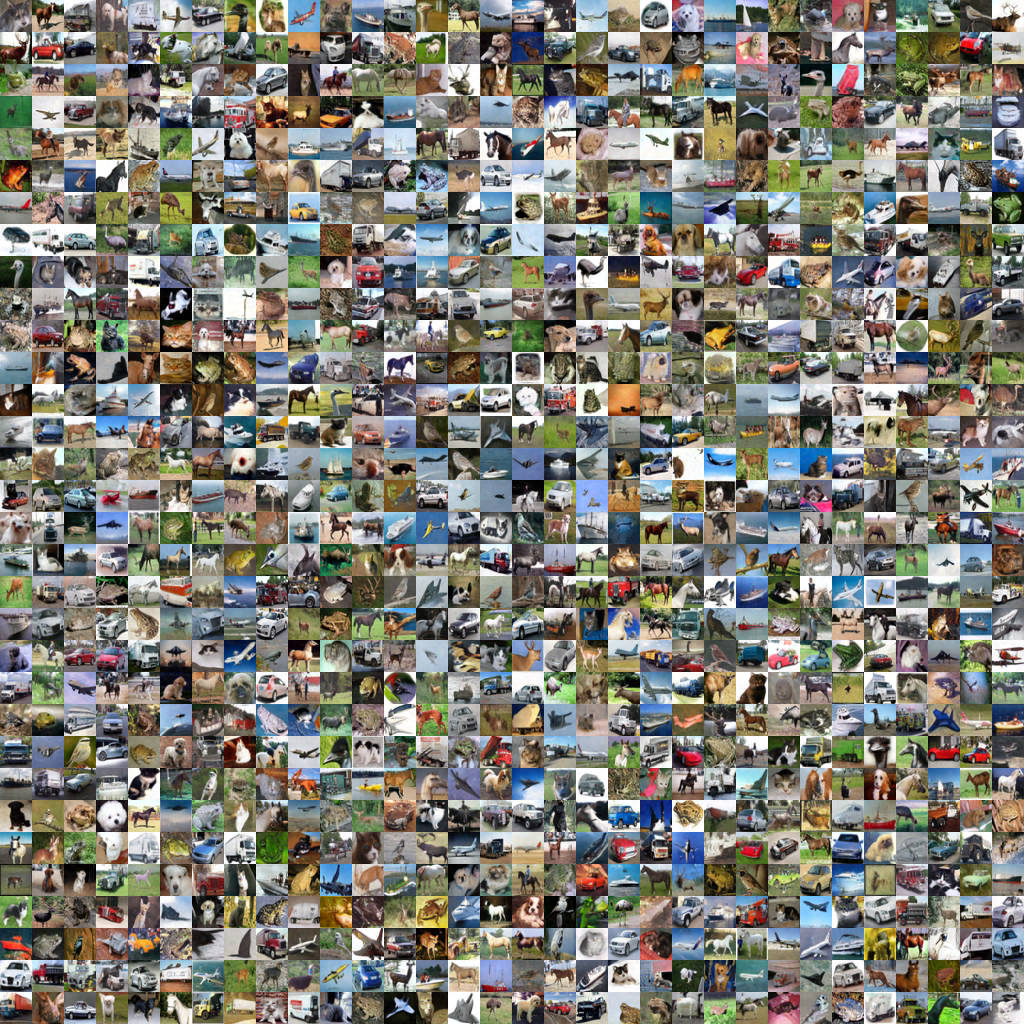}
    \caption{Unconditional CIFAR-10 $32\times32$ random images generated with DI+EL (FID: 3.95).}
\end{figure}

\begin{figure}[p]
    \centering
    \includegraphics[width=0.9\textwidth]{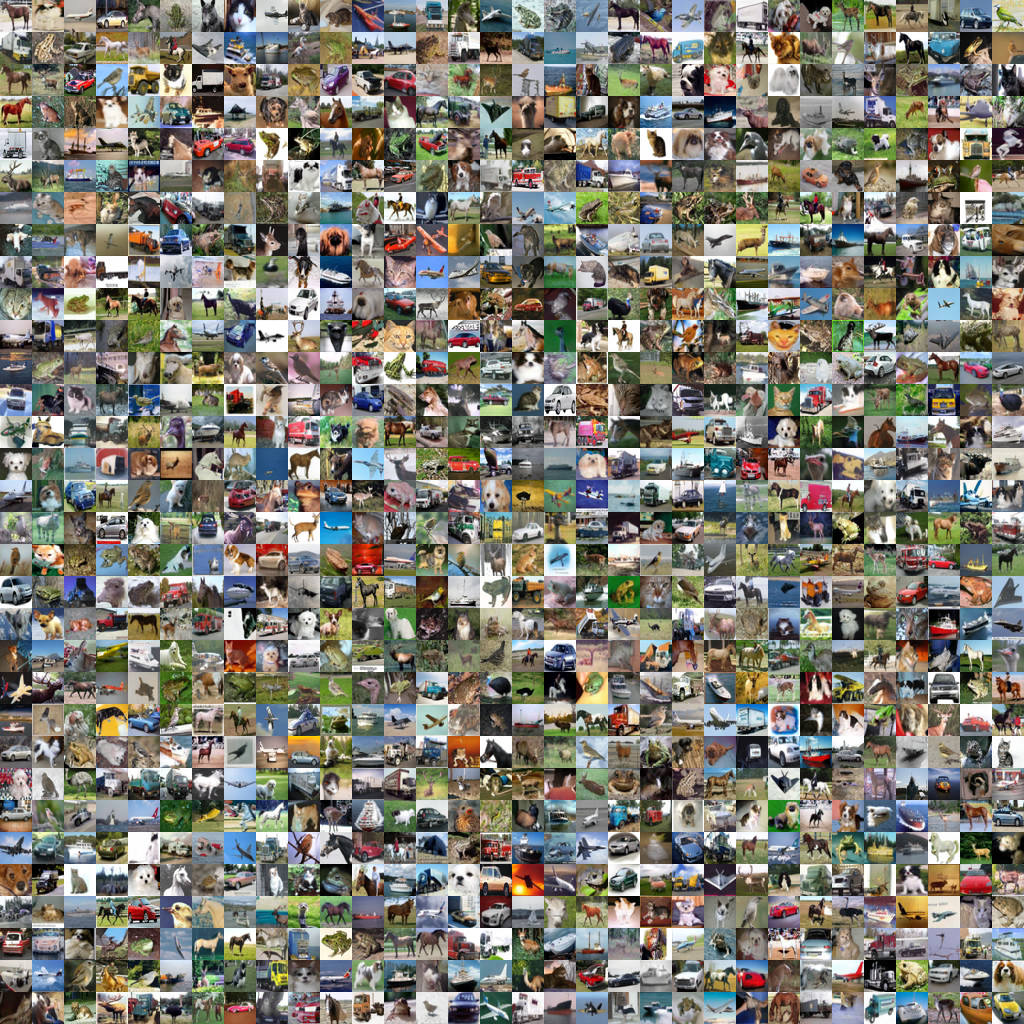}
    \caption{Unconditional CIFAR-10 $32\times32$ random images generated with SiD\textsuperscript{2}A+EL (FID: 1.475).}
    \label{fig:cifar10_SID_UNCON}
\end{figure}

\begin{figure}[p]
    \centering
    \includegraphics[width=0.9\textwidth]{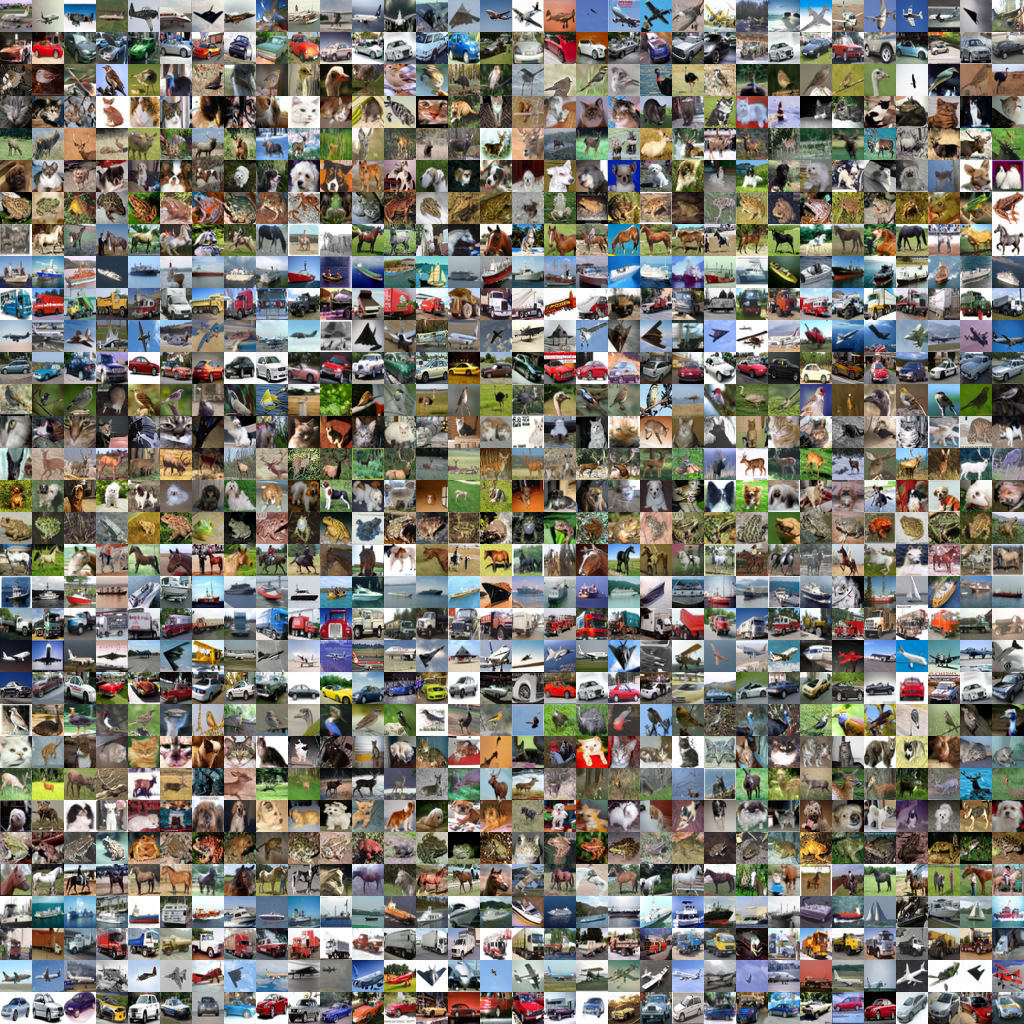}
    \caption{Label-conditioned CIFAR-10 $32\times32$ random images generated with SiD\textsuperscript{2}A+EL (FID: 1.38).}
    \label{fig:cifar10_SID_CON}
\end{figure}

\begin{figure}[p]
\label{FFHQ 64 × 64 photo}
    \centering
    \includegraphics[width=0.9\textwidth]{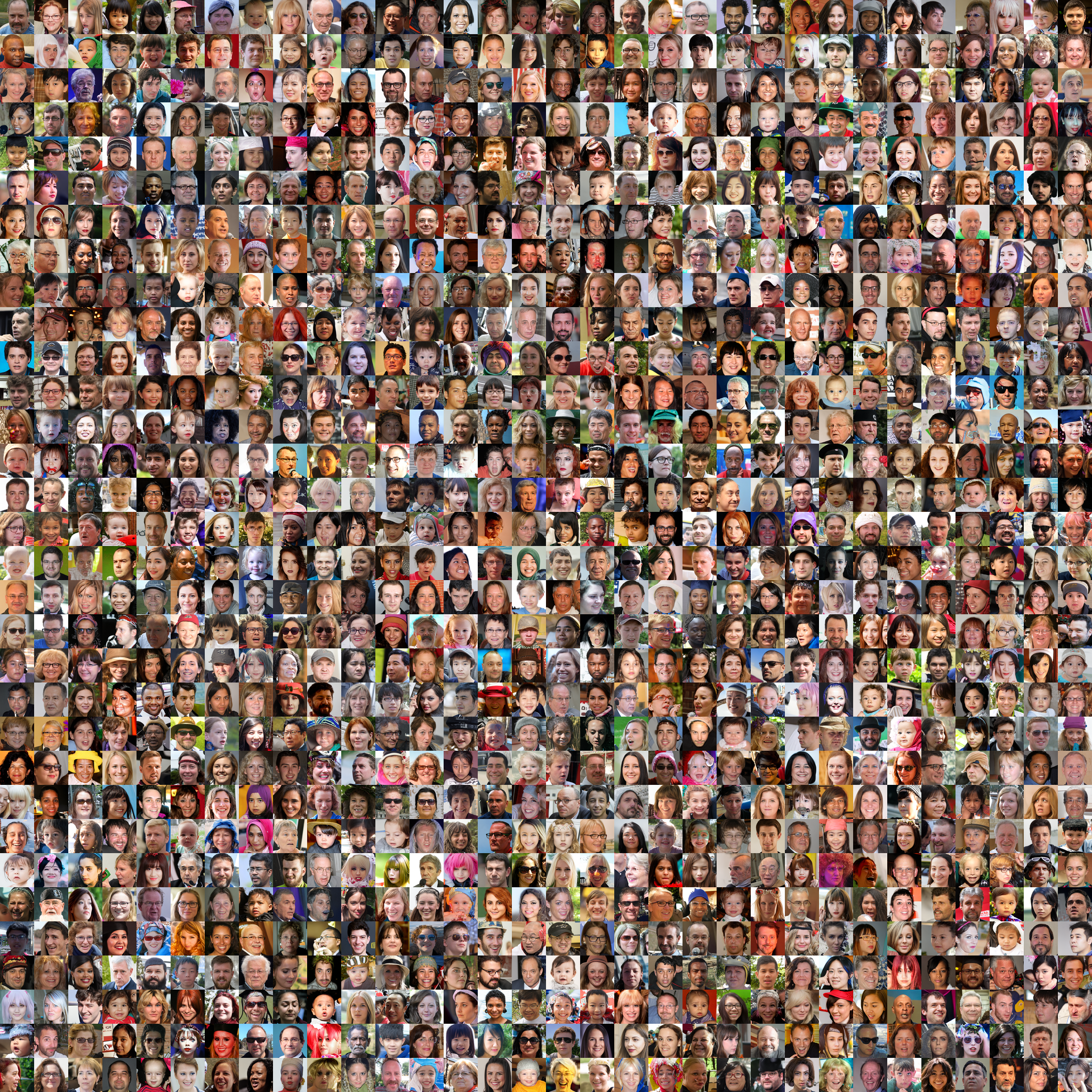}
    \caption{FFHQ $64\times64$ random images generated with SiD\textsuperscript{2}A+EL (FID: 1.06).}
\end{figure}

\begin{figure}[p]
\label{AFHQ 64 × 64 photo}
    \centering
    \includegraphics[width=0.9\textwidth]{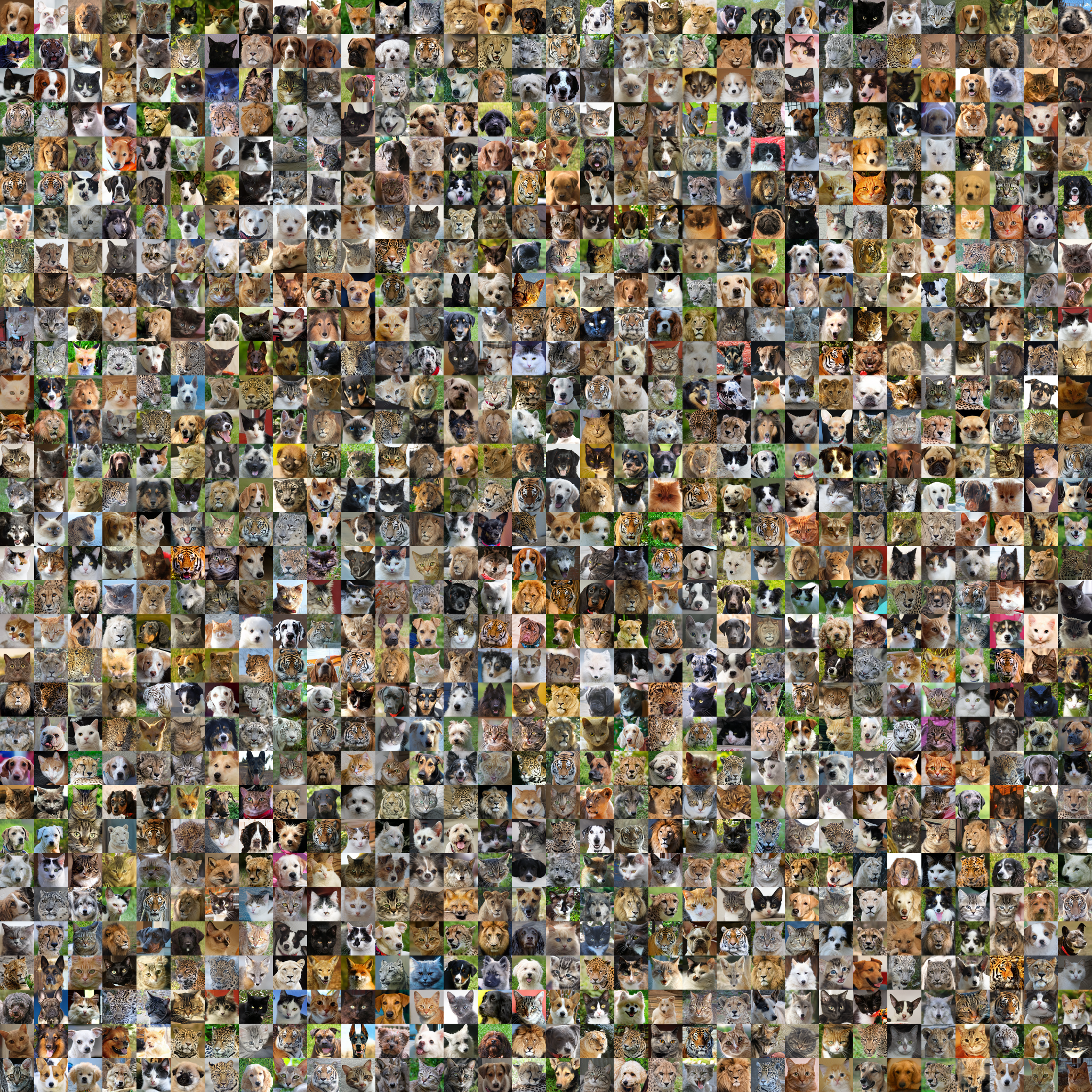}
    \caption{AFHQ-V2 $64\times64$ random images generated with SiD\textsuperscript{2}A+EL (FID: 1.26).}
\end{figure}

\begin{figure}[p]
\label{Image 512 × 512 photo}
    \centering
    \includegraphics[width=0.9\textwidth]{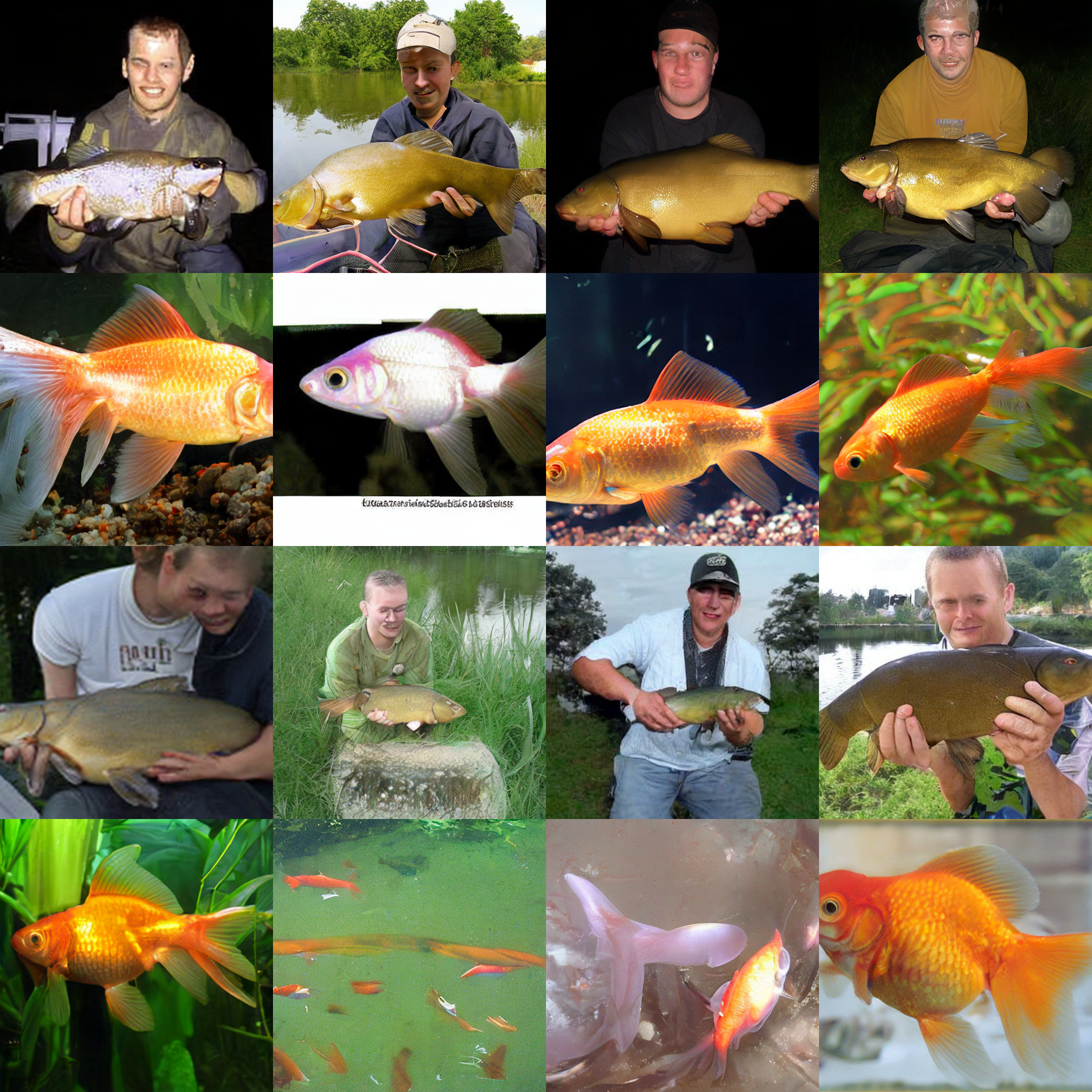}
    \caption{ImageNet $512\times512$ random images generated with SiD\textsuperscript{2}A+EL (FID: 2.132).}
\end{figure}

\FloatBarrier

\end{document}